\documentclass[journal]{IEEEtran}

\usepackage{amssymb, amsmath, amsthm, amsfonts, enumerate}
\usepackage{graphicx}
\usepackage{url}
\usepackage{pdfpages}
\usepackage{color}
\usepackage{blindtext}
\usepackage{ragged2e}
\usepackage[utf8]{inputenc}
\usepackage{subcaption}
\usepackage{glossaries}
\usepackage{nomencl}
\usepackage{fancyhdr}
\usepackage{titlesec}
\usepackage{mathtools}
\usepackage{dirtytalk}
\usepackage{booktabs}
\usepackage{caption}
\usepackage{siunitx}
\usepackage[T1]{fontenc}
\usepackage{epstopdf}
\usepackage{commath}
\usepackage{float}
\usepackage{setspace}
\usepackage{cite}
\usepackage{algorithmic}
\usepackage{algorithm}
\usepackage{stfloats}
\usepackage{placeins}
\usepackage{multirow}
\usepackage{tikz}
\usepackage{tabularx}
\usepackage{xcolor}

\usetikzlibrary{shapes.geometric, arrows.meta, positioning, fit, backgrounds, shadows}

\definecolor{grayFill}   {RGB}{241,239,232}
\definecolor{grayLine}   {RGB}{95,94,90}
\definecolor{grayText}   {RGB}{68,68,65}
\definecolor{purpleFill} {RGB}{238,237,254}
\definecolor{purpleLine} {RGB}{83,74,183}
\definecolor{purpleText} {RGB}{60,52,137}
\definecolor{redFill}    {RGB}{252,235,235}
\definecolor{redLine}    {RGB}{163,45,45}
\definecolor{redText}    {RGB}{121,31,31}
\definecolor{amberFill}  {RGB}{250,238,218}
\definecolor{amberLine}  {RGB}{133,79,11}
\definecolor{amberText}  {RGB}{99,56,6}
\definecolor{limeFill}   {RGB}{234,243,222}
\definecolor{limeLine}   {RGB}{59,109,17}
\definecolor{limeText}   {RGB}{39,80,10}
\definecolor{pinkFill}   {RGB}{251,234,240}
\definecolor{pinkLine}   {RGB}{153,53,86}
\definecolor{pinkHdr}    {RGB}{114,36,62}
\definecolor{burnFill}   {RGB}{250,236,231}
\definecolor{burnLine}   {RGB}{153,60,29}
\definecolor{burnHdr}    {RGB}{113,43,19}
\definecolor{blueFill}   {RGB}{230,241,251}
\definecolor{blueLine}   {RGB}{24,95,165}
\definecolor{blueHdr}    {RGB}{12,68,124}
\definecolor{tealFill}   {RGB}{225,245,238}
\definecolor{tealLine}   {RGB}{15,110,86}
\definecolor{tealHdr}    {RGB}{8,80,65}
\definecolor{beigeF}     {RGB}{245,244,237}
\definecolor{beigeL}     {RGB}{165,164,160}
\definecolor{arrowClr}   {RGB}{115,114,108}
\definecolor{greenArr}   {RGB}{99,153,34}
\definecolor{titleClr}   {RGB}{20,20,19}
\definecolor{bodyClr}    {RGB}{61,61,58}

\theoremstyle{definition}

\usepackage[english]{babel}

\begin{document}

\title{Prompt-to-Paper: Agentic AI System for Bioinformatics}

\author{Ramsha~Kamran, Maheera Amjad,
        Zartasha~Mustansar,
        Arsalan~Shaukat,
        Salma~Sherbaz,
        and Muhammad~U.~S.~Khan%
\thanks{R. Kamran, M. Amjad, Z. Mustansar, S. Sherbaz, and M. U. S. Khan are with the School of Interdisciplinary Engineering and Sciences (SINES), National University of Sciences and Technology (NUST), Islamabad, Pakistan.}%
\thanks{A. Shaukat is with the College of Electrical and Mechanical Engineering (CEME), NUST, Rawalpindi, Pakistan.}}

\markboth{Preprint --- Prompt-to-Paper: Agentic AI System for Bioinformatics}%
{Kamran \MakeLowercase{\textit{et al.}}: Prompt-to-Paper: Agentic AI System for Bioinformatics}

\maketitle
\thispagestyle{plain}

\begin{abstract}
 While recent advances in large language models have enabled end-to-end automated manuscript generation, existing systems suffer from three critical deficiencies: (i) generated claims are not deterministically grounded in verifiable literature, (ii) experimental results are frequently fabricated rather than executed, and (iii) there exists no standardized, multi-dimensional framework to assess whether AI-generated manuscripts meet the quality and rigor required for real-world publication. We present \textit{Prompt-to-Paper}, a multi-stage multi-agent framework that directly addresses this evaluation gap through three integrated innovations. First, a deterministic retrieval-augmented generation pipeline with section-aware relevance scoring and snowball citation expansion grounds every claim in a verifiable corpus of 60--100 papers. Second, an autonomous coding agent executes real computational biology experiments, replacing synthetic outputs with genuine numerical results. Third, an \textbf{eight-dimensional} automated quality scorer, benchmarked with approximate reference statistics from published papers and augmented with explicit hallucination penalties, provides standardized, reproducible quality assessments. The quality-driven improvement loop uses a context-rich reviser that routes each iteration to one of three researcher actions (adding statistical analysis from executed results, gathering new literature evidence via a RAG store, or rewriting for clarity) and fires a deep research cycle every ten iterations to re-run experiments and re-manuscript from stronger outputs. We validate the system on five bioinformatics case studies; all five cases compiled submission-formatted PDFs with zero out-of-range citations. The improvement loop raises manuscript quality by an average of $+17.96$ points on a 0--100 scale (maximum $+26.04$). As partial external checks, three independent large language models converged on the same quality ranking, and a human reviewer scored the five manuscripts at an average of 7.0 out of 10. Complete manuscripts are produced at approximately \$0.31 per paper. 
\end{abstract}

\begin{IEEEkeywords}
agentic AI, bioinformatics, automated manuscript generation, retrieval-augmented generation, scientific quality assessment, deep research cycles.
\end{IEEEkeywords}

\IEEEpeerreviewmaketitle

\section{Introduction}
\label{sec:introduction}

The human capacity to keep up with the scientific literature has been surpassed. The number of publications is growing, and researchers are no longer able to track the field~\cite{Landhuis2016}. This quantitative pressure has contributed to a measurable decrease in quality and persistent problems of reproducibility and validation~\cite{Ioannidis2005, Baker2016}. As retractions and corrections have increased, the scientific system faces an integrity crisis it can no longer reliably self-correct~\cite{Brainard2018, RetractionWatch}.

AI has transformed the process of scientific writing, yet this shift introduces a challenge that remains unresolved: how do we validate automated papers when researchers already struggle to validate human-written ones? The real question is not whether AI can write a paper, but whether the paper it writes is actually good and fills a genuine scientific void.

The impacts are beginning to be felt. A paper published in February 2024 on spermatogonial stem cells passed the review process despite concerns about AI-generated content and was subsequently retracted~\cite{Guo2024Retracted}. Similarly, Sakana AI's system for automating idea generation, experiments, paper writing, and peer review passed the initial review stages of a real workshop, yet the authors noted hallucinated citations, weak novelty, experimental errors, and poor reproducibility~\cite{Lu2026, Zhu2025}. There is a common cause in both cases: evaluating the quality, validity, and publishability of AI-generated scientific papers is still a major unsolved challenge~\cite{Hopner2025, Zhu2025}.

We present \textit{Prompt-to-Paper: An Agentic AI System for Bioinformatics}, a multi-agent framework that closes this gap through:
\begin{itemize}
  \item Deterministic RAG-based retrieval grounded in 60--100 verified real papers,
  \item An autonomous coding agent that executes real experiments and injects verified numeric results into every manuscript section,
  \item An eight-dimensional automated quality scorer with hallucination penalties,
  \item A context-rich iterative improvement loop with deep research cycles that sustain quality gains past the prose-polishing plateau.
\end{itemize}

\section{Related Work}
\label{sec:related}

\subsection{Knowledge-Grounded Scientific Writing}
PaperRobot~\cite{PaperRobot1} constructs a knowledge graph of over 1.6 million biomedical entities and uses graph and text attention to predict novel hypotheses, generating abstracts, conclusions, and follow-on titles. In Turing tests its abstracts were preferred over human ones 30\% of the time. However, the system uses a static, pre-built knowledge graph; it does not retrieve or verify facts from live literature during generation, and never executes real experiments. PaSa~\cite{PaSa9} addresses the retrieval problem with an LLM-powered search agent that autonomously traverses citation networks, achieving significant recall improvements. ScholarGym~\cite{ScholarGym30} provides a 570K-paper simulation environment for reproducible evaluation of research workflows. These systems focus on retrieval or evaluation but lack an integrated writing or revision component.

\subsection{Iterative Manuscript Refinement}
CycleResearcher~\cite{CycleResearcher2} trains a policy model that generates complete papers and a reward model (CycleReviewer) that mimics peer review, both updated iteratively via reinforcement learning. CycleReviewer reduces reviewer-score prediction error by 27\% compared to individual human reviewers, and the generated papers achieve a mean simulated score of 5.36/10, close to the human-preprint average of 5.24. However, the authors explicitly report that all experimental numbers are synthetic rather than executed, the system is locked to machine learning papers, and the evaluation relies entirely on the simulated reviewer with no ground-truth quality check. OpenLens AI~\cite{OpenLens12} addresses health informatics through multi-agent literature review and data analysis, but lacks rigorous benchmarking against general-purpose systems and does not execute real experiments.

\subsection{Research Gaps}
The above analysis reveals three open challenges. First, current systems rely on domain-specific training data or static knowledge graphs and do not generalise. Second, deterministic retrieval over a verifiable live corpus is absent, and hallucinated or synthetic results are common. Third, no system combines real experiment execution with a multi-dimensional, penalty-equipped quality scorer. Our system addresses all three.

\section{Methodology}
\label{sec:methodology}

Research Landscape Explorer v4 (RLEv4), the implementation backbone of \textit{Prompt-to-Paper}, is a multi-agent pipeline that converts a research topic into a submission-formatted manuscript with minimal human involvement. It overcomes three principal weaknesses of prior systems: lack of verifiable literature grounding, absence of real experiment execution, and no iterative quality control. The pipeline has five high-level stages: literature acquisition and relevance scoring, knowledge graph construction and claim alignment, manuscript generation, autonomous bioinformatics experimentation, and context-rich quality-driven revision.

While systems like Gemini~\cite{woodruff2026accelerating} demonstrate AI as a collaborative partner in mathematical discovery, RLEv4 provides a fully automated, production-ready pipeline for the entire research lifecycle without requiring expert-level prompting at every stage.

\begin{figure}[t]
\centering
\includegraphics[width=\columnwidth]{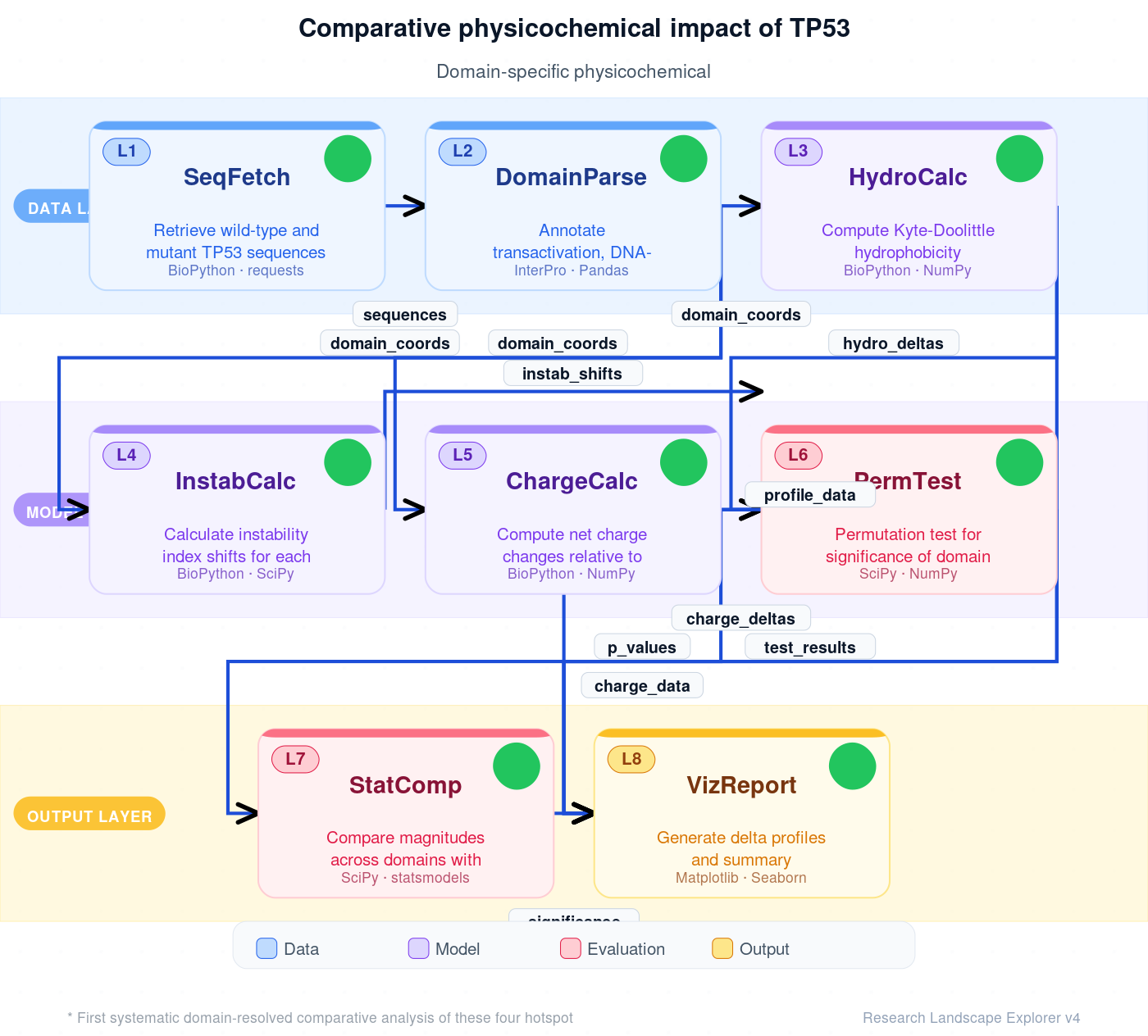}
\caption{Planning-agent output for the TP53 hotspot mutation physicochemical analysis. Each block is a functional stage grouped into data, model, evaluation, and output layers; arrows trace the data flow from raw sequences through physicochemical computation and permutation testing to the final visualisation.}
\label{fig:architecture}
\end{figure}

\subsection{Literature Acquisition and Relevance Scoring}

Given a research topic $t$, RLEv4 retrieves real academic papers from Semantic Scholar and, when configured, Tavily. For each paper $p$, the pipeline extracts its title, abstract, year, citation count, venue, and, when available via PDF acquisition and section segmentation, full-text sections. A section-aware relevance score is computed as:
\begin{equation}
  R(p,t) = \sum_{s \in S} w_s \cdot \cos\!\left(\mathbf{e}_s(p),\, \mathbf{e}_q(t)\right),
  \label{eq:relevance}
\end{equation}
where $S$ is the set of parsed sections (abstract, methods, results, introduction, discussion, conclusion); $w_s$ are empirically set weights (abstract 0.30, methods 0.25, results 0.20, introduction 0.12, discussion 0.08, conclusion 0.05); $\mathbf{e}_s(p)$ is a SPECTER2 embedding of section $s$ (or a BM25+ salient-section fallback when SPECTER2 is unavailable); and $\mathbf{e}_q(t)$ is the query embedding.

An initial seed set of up to 60 papers is expanded by snowball sampling inspired by PaSa~\cite{PaSa9}: for each seed paper the pipeline fetches its references and citations via the Semantic Scholar batch API, computes $R(\cdot,t)$ for candidates, and retains those exceeding a relevance threshold $\tau \in [0.05, 0.08]$. Two snowball iterations produce a final corpus of 60--100 papers. A diversity ranker then orders the corpus by a composite score combining citation count, recency, and automatically detected research stance (favour/against/comparison), using the heuristic keyword approach from EpidemIQs~\cite{EpidemIQs14} to surface contradictory evidence alongside supportive work.

\subsection{Knowledge Graph and Claim Alignment}

From the top-ranked papers the pipeline constructs a directed graph $G=(V,E)$ in which nodes represent papers and edges represent citations. PageRank scores $\mathrm{PR}(v)$ are computed over the paper subgraph to identify influential works. In parallel, a claim extraction module prompts the leader model with each paper's abstract, results, and conclusion to produce up to five falsifiable claims per paper, stored in a relational SQLite database alongside the corpus.

\begin{figure*}[t]
\centering
\includegraphics[width=\textwidth]{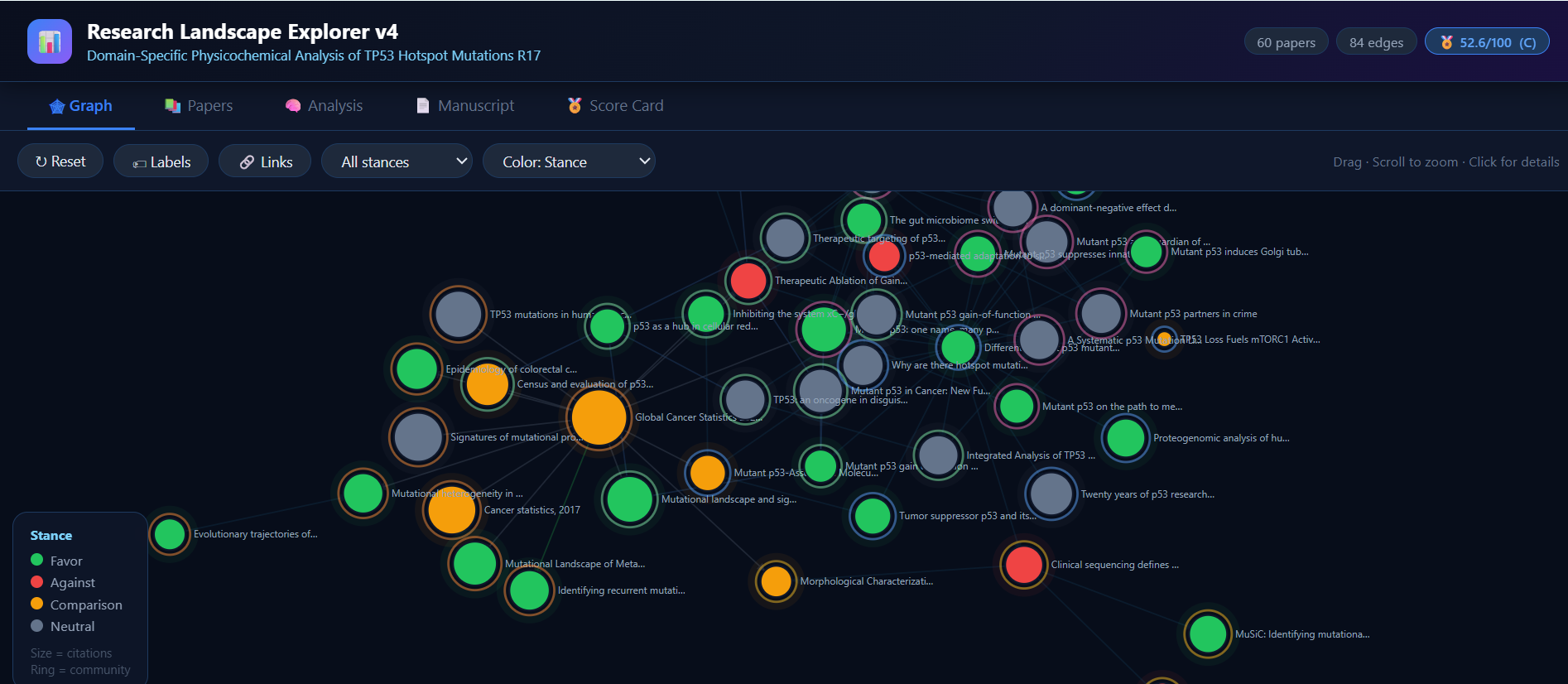}
\caption{Knowledge graph of claim alignments for the TP53 query. Nodes are papers, sized by PageRank; edges are coloured by claim relation: \textsc{Supports} (green), \textsc{Contradicts} (red), \textsc{Orthogonal} (blue). Red edges mark genuine disagreements that the manuscript's gap analysis targets directly.}
\label{fig:knowledge_graph}
\end{figure*}

\begin{figure*}[t]
\centering
\includegraphics[width=\textwidth]{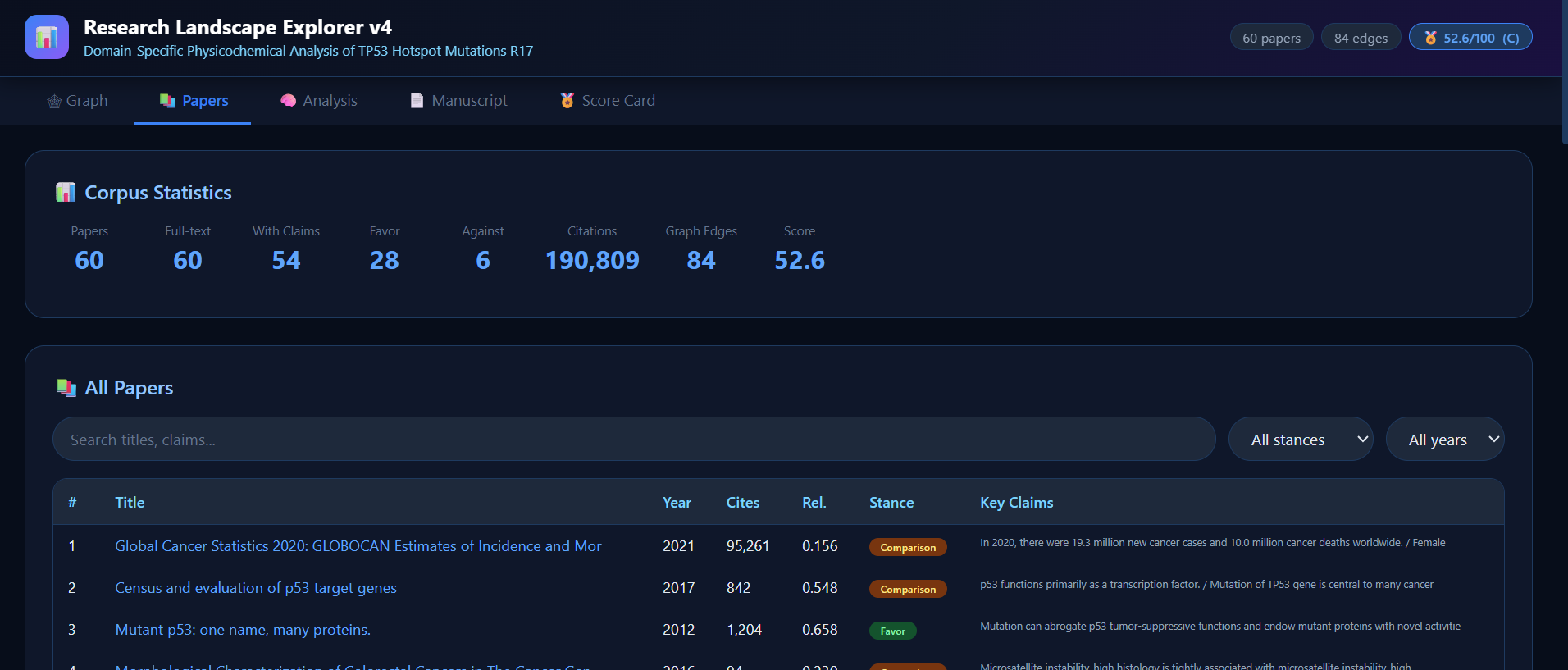}
\caption{Papers view of the RLEv4 interactive dashboard for the TP53 hotspot-mutation query, showing corpus statistics, per-paper citation counts, relevance scores, detected research stances, and extracted key claims.}
\label{fig:dashboard_papers}
\end{figure*}

Claim alignment builds a matrix of pairwise relations. For claims $c_i$ and $c_j$ from different papers, the leader model classifies the relation as \textsc{Supports}, \textsc{Contradicts}, or \textsc{Orthogonal}; a heuristic keyword-overlap fallback is used when the LLM is unavailable. The resulting contradictions and consensus statements directly inform the manuscript's gap analysis, extending the pairwise comparison paradigm of CycleResearcher~\cite{CycleResearcher2} to a live, corpus-grounded setting.

\subsection{Leader-Worker Model Routing}
\label{ssec:llm_routing}

All LLM calls are routed through a two-tier DeepSeek-only stack enforced by the hard switch \texttt{USE\_ONLY\_DEEPSEEK=1}. The \textit{leader} role (quality judging, gap-finding, synthesis, and deep-cycle re-manuscripting) uses \textit{deepseek-v4-pro}, a reasoning-capable model whose chain-of-thought aids analytical depth. The \textit{worker} role (section writing, code generation, JSON planning) uses \textit{deepseek-chat}, a fast non-reasoning model that reliably emits structured outputs without leaking planning prose into the manuscript. This split is the direct result of a failure observed with earlier cross-model configurations: when the quality judge spent its output budget on hidden reasoning rather than emitting clean JSON, the G-Eval scorer silently fell back to a neutral prior for every dimension, making the improvement loop content-independent and causing the overall score to freeze regardless of manuscript quality. Routing the judge to a reasoning model while keeping section writing on a fast chat model resolves both problems simultaneously.

The leader constructs a JSON specification of the pipeline's functional layers and their data flow, as shown in the planning-agent output in Fig.~\ref{fig:architecture}, before any writing begins. The worker then drafts each section in sequence, guided by structured prompts that enforce active voice, precise numerical reporting, and avoidance of generic filler phrases. After generation the leader re-evaluates each section, providing revision feedback before the worker revises.

\subsection{Autonomous Bioinformatics Experimentation}
\label{ssec:coding}

For computational biology topics, RLEv4 invokes an autonomous coding agent inspired by the execution layer of EpidemIQs~\cite{EpidemIQs14}. The agent receives a detailed task query embedding the bioinformatics problem, hardcoded reference data, and all required algorithmic steps, then autonomously writes, executes, and debugs Python scripts that:
\begin{itemize}
  \item Use hardcoded sequences and matrices (no network calls during experiments), with NCBI Entrez fallbacks wrapped in five-retry exponential-backoff loops.
  \item Perform progressive multiple sequence alignment (match $=2$, mismatch $=-1$, gap open $=-5$, gap extend $=-2$) and compute Jukes-Cantor distances $d = -\frac{3}{4}\ln\!\left(1 - \frac{4p}{3}\right)$.
  \item Execute statistical validation via an injected \texttt{rigor.py} module: permutation tests ($n=5{,}000$ shuffles), bootstrap confidence intervals, Benjamini-Hochberg multiple-comparison correction, Cohen's $d$ effect sizes, and null-baseline comparisons.
  \item Save all numeric findings to a canonical \texttt{results.json} and at least one publication-quality \texttt{plot.png}.
\end{itemize}

A pre-execution syntax gate (\texttt{ast.parse}) catches unclosed brackets and truncated f-strings before they consume a full attempt; on failure a focused repair prompt targeting the exact offending line is sent before falling back to a full regeneration. The agent runs in a sandboxed environment with a one-hour timeout and up to 12 genuine execution attempts (\texttt{BIO\_MAX\_REAL\_ATTEMPTS=12}). Successful runs produce a canonical \texttt{CanonicalResults} object whose numbers are injected verbatim into every manuscript section, guaranteeing numerical consistency across the paper. Unlike CycleResearcher~\cite{CycleResearcher2}, which explicitly fabricates experimental results, our agent executes real computations and the manuscript reports only those verified outputs.

\subsection{Quality Scoring}
\label{ssec:scoring}

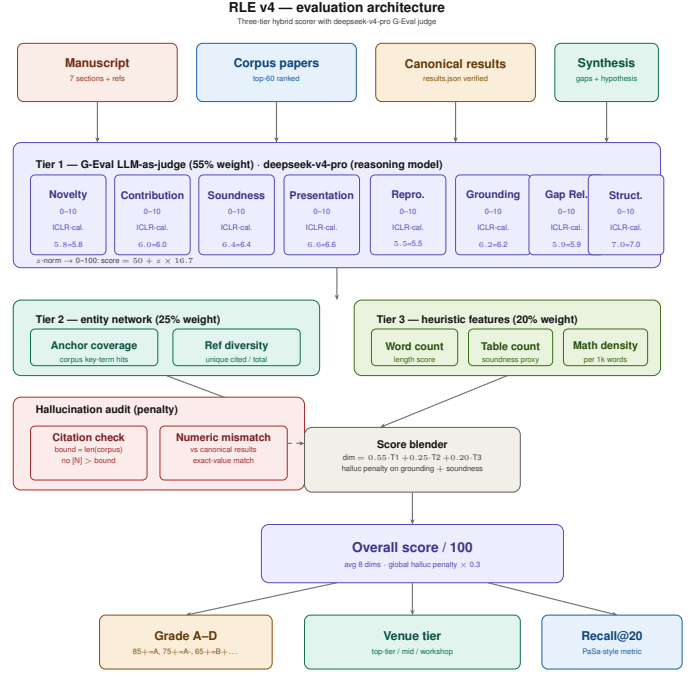
\begin{figure}[tbp]
\centering
\definecolor{purpleHdr}   {RGB}{60,52,137}
\definecolor{purpleSub}   {RGB}{83,74,183}
\tikzset{
  arr/.style  = {-{Stealth[length=3.5pt,width=2.8pt]}, draw=arrowClr, line width=0.75pt},
  darr/.style = {arr, dashed},
}
\resizebox{\columnwidth}{!}{%
\begin{tikzpicture}[font=\sffamily]
\node[font=\sffamily\small\bfseries, text=titleClr] at (8.5,18.50) {RLE v4 --- evaluation architecture};
\node[font=\sffamily\tiny, text=bodyClr] at (8.5,18.15) {Three-tier hybrid scorer with deepseek-v4-pro G-Eval judge};
\draw[fill=burnFill,draw=burnLine,line width=0.4pt,rounded corners=4pt] (1.10,16.30) rectangle (4.80,17.65);
\node[font=\sffamily\footnotesize\bfseries,text=burnHdr] at (2.95,17.18) {Manuscript};
\node[font=\sffamily\tiny,text=burnLine] at (2.95,16.83) {7 sections + refs};
\draw[fill=blueFill,draw=blueLine,line width=0.4pt,rounded corners=4pt] (5.25,16.30) rectangle (8.95,17.65);
\node[font=\sffamily\footnotesize\bfseries,text=blueHdr] at (7.10,17.18) {Corpus papers};
\node[font=\sffamily\tiny,text=blueLine] at (7.10,16.83) {top-60 ranked};
\draw[fill=amberFill,draw=amberLine,line width=0.4pt,rounded corners=4pt] (9.40,16.30) rectangle (13.10,17.65);
\node[font=\sffamily\footnotesize\bfseries,text=amberText] at (11.25,17.18) {Canonical results};
\node[font=\sffamily\tiny,text=amberLine] at (11.25,16.83) {results.json verified};
\draw[fill=tealFill,draw=tealLine,line width=0.4pt,rounded corners=4pt] (13.55,16.30) rectangle (15.95,17.65);
\node[font=\sffamily\footnotesize\bfseries,text=tealHdr] at (14.75,17.18) {Synthesis};
\node[font=\sffamily\tiny,text=tealLine] at (14.75,16.83) {gaps + hypothesis};
\draw[arr](2.95,16.30)--(2.95,15.35); \draw[arr](7.10,16.30)--(7.10,15.35);
\draw[arr](11.25,16.30)--(11.25,15.35); \draw[arr](14.75,16.30)--(14.75,15.35);
\draw[fill=purpleFill,draw=purpleLine,line width=0.4pt,rounded corners=5pt] (1.00,12.45) rectangle (16.00,15.35);
\node[font=\sffamily\scriptsize\bfseries,text=titleClr,anchor=west] at (1.40,14.83)
  {Tier 1 --- G-Eval LLM-as-judge (55\% weight) $\cdot$ deepseek-v4-pro (reasoning model)};
\foreach \x/\lbl/\mu in {
  2.275/Novelty/5.8, 4.225/Contribution/6.0, 6.175/Soundness/6.4, 8.150/Presentation/6.6,
  10.15/Repro./5.5, 12.125/Grounding/6.2, 13.825/Gap Rel./5.9, 15.20/Struct./7.0}{
  \pgfmathsetmacro\xl{\x-0.875}
  \pgfmathsetmacro\xr{\x+0.875}
  \draw[fill=purpleFill,draw=purpleLine,line width=0.4pt,rounded corners=3pt](\xl,12.75)rectangle(\xr,14.60);
  \node[font=\sffamily\scriptsize\bfseries,text=purpleHdr] at (\x,14.12) {\lbl};
  \node[font=\sffamily\tiny,text=purpleSub] at (\x,13.75) {0--10};
  \node[font=\sffamily\tiny,text=purpleSub] at (\x,13.38) {ICLR-cal.};
  \node[font=\sffamily\tiny,text=purpleSub] at (\x,13.00) {$\mu$=\mu};
}
\node[font=\sffamily\tiny,text=bodyClr,anchor=west] at (1.40,12.60) {$z$-norm $\to$ 0--100: score $= 50 + z\times 16.7$};
\draw[arr](8.50,12.45)--(8.50,11.70);
\draw[fill=tealFill,draw=tealLine,line width=0.4pt,rounded corners=5pt] (1.00,9.95) rectangle (8.10,11.70);
\node[font=\sffamily\scriptsize\bfseries,text=titleClr,anchor=west] at (1.40,11.23) {Tier 2 --- entity network (25\% weight)};
\draw[fill=tealFill,draw=tealLine,line width=0.4pt,rounded corners=3pt](1.40,10.18)rectangle(4.35,11.03);
\node[font=\sffamily\scriptsize\bfseries,text=tealHdr] at (2.875,10.63) {Anchor coverage};
\node[font=\sffamily\tiny,text=tealLine] at (2.875,10.30) {corpus key-term hits};
\draw[fill=tealFill,draw=tealLine,line width=0.4pt,rounded corners=3pt](4.70,10.18)rectangle(7.65,11.03);
\node[font=\sffamily\scriptsize\bfseries,text=tealHdr] at (6.175,10.63) {Ref diversity};
\node[font=\sffamily\tiny,text=tealLine] at (6.175,10.30) {unique cited / total};
\draw[fill=limeFill,draw=limeLine,line width=0.4pt,rounded corners=5pt] (8.90,9.95) rectangle (16.00,11.70);
\node[font=\sffamily\scriptsize\bfseries,text=titleClr,anchor=west] at (9.30,11.23) {Tier 3 --- heuristic features (20\% weight)};
\draw[fill=limeFill,draw=limeLine,line width=0.4pt,rounded corners=3pt](9.30,10.18)rectangle(11.30,11.03);
\node[font=\sffamily\scriptsize\bfseries,text=limeText] at (10.30,10.63) {Word count};
\node[font=\sffamily\tiny,text=limeLine] at (10.30,10.30) {length score};
\draw[fill=limeFill,draw=limeLine,line width=0.4pt,rounded corners=3pt](11.55,10.18)rectangle(13.50,11.03);
\node[font=\sffamily\scriptsize\bfseries,text=limeText] at (12.525,10.63) {Table count};
\node[font=\sffamily\tiny,text=limeLine] at (12.525,10.30) {soundness proxy};
\draw[fill=limeFill,draw=limeLine,line width=0.4pt,rounded corners=3pt](13.75,10.18)rectangle(15.70,11.03);
\node[font=\sffamily\scriptsize\bfseries,text=limeText] at (14.725,10.63) {Math density};
\node[font=\sffamily\tiny,text=limeLine] at (14.725,10.30) {per 1k words};
\draw[arr](4.55,9.95)--(7.75,8.75); \draw[arr](12.45,9.95)--(9.50,8.75);
\draw[fill=redFill,draw=redLine,line width=0.4pt,rounded corners=5pt](1.00,7.30)rectangle(7.75,9.45);
\node[font=\sffamily\scriptsize\bfseries,text=titleClr,anchor=west] at (1.40,9.15) {Hallucination audit (penalty)};
\draw[fill=redFill,draw=redLine,line width=0.4pt,rounded corners=3pt](1.40,7.53)rectangle(4.10,8.78);
\node[font=\sffamily\scriptsize\bfseries,text=redText] at (2.750,8.53) {Citation check};
\node[font=\sffamily\tiny,text=redLine] at (2.750,8.23) {bound = len(corpus)};
\node[font=\sffamily\tiny,text=redLine] at (2.750,7.98) {no [N] $>$ bound};
\draw[fill=redFill,draw=redLine,line width=0.4pt,rounded corners=3pt](4.40,7.53)rectangle(7.35,8.78);
\node[font=\sffamily\scriptsize\bfseries,text=redText] at (5.875,8.53) {Numeric mismatch};
\node[font=\sffamily\tiny,text=redLine] at (5.875,8.23) {vs canonical results};
\node[font=\sffamily\tiny,text=redLine] at (5.875,7.98) {exact-value match};
\draw[darr](7.35,8.38)--(7.75,8.38);
\draw[fill=grayFill,draw=grayLine,line width=0.4pt,rounded corners=5pt](7.75,7.25)rectangle(12.75,8.75);
\node[font=\sffamily\scriptsize\bfseries,text=titleClr] at (10.25,8.35) {Score blender};
\node[font=\sffamily\tiny,text=bodyClr] at (10.25,8.05) {dim $= 0.55\cdot$T1 $+ 0.25\cdot$T2 $+ 0.20\cdot$T3};
\node[font=\sffamily\tiny,text=bodyClr] at (10.25,7.78) {halluc penalty on grounding $+$ soundness};
\draw[arr](10.25,7.25)--(10.25,6.50);
\draw[fill=purpleFill,draw=purpleLine,line width=0.4pt,rounded corners=5pt](6.75,5.15)rectangle(13.75,6.50);
\node[font=\sffamily\small\bfseries,text=purpleHdr] at (10.25,5.98) {Overall score / 100};
\node[font=\sffamily\tiny,text=purpleSub] at (10.25,5.55) {avg 8 dims $\cdot$ global halluc penalty $\times$ 0.3};
\draw[arr](8.50,5.15)--(5.00,4.40); \draw[arr](10.25,5.15)--(10.25,4.40); \draw[arr](12.00,5.15)--(15.00,4.40);
\draw[fill=amberFill,draw=amberLine,line width=0.4pt,rounded corners=4pt](3.00,3.15)rectangle(7.00,4.40);
\node[font=\sffamily\footnotesize\bfseries,text=amberText] at (5.00,3.93) {Grade A--D};
\node[font=\sffamily\tiny,text=amberLine] at (5.00,3.55) {85$+$=A, 75$+$=A-, 65$+$=B$+$\ldots};
\draw[fill=tealFill,draw=tealLine,line width=0.4pt,rounded corners=4pt](7.75,3.15)rectangle(12.75,4.40);
\node[font=\sffamily\footnotesize\bfseries,text=tealHdr] at (10.25,3.93) {Venue tier};
\node[font=\sffamily\tiny,text=tealLine] at (10.25,3.55) {top-tier / mid / workshop};
\draw[fill=blueFill,draw=blueLine,line width=0.4pt,rounded corners=4pt](13.25,3.15)rectangle(16.50,4.40);
\node[font=\sffamily\footnotesize\bfseries,text=blueHdr] at (14.875,3.93) {Recall@20};
\node[font=\sffamily\tiny,text=blueLine] at (14.875,3.55) {PaSa-style metric};
\end{tikzpicture}}
\caption{RLEv4 evaluation architecture. The three-tier hybrid scorer combines deepseek-v4-pro G-Eval judgement (Tier 1, 55\%), corpus-entity network metrics (Tier 2, 25\%), and heuristic surface features (Tier 3, 20\%), with a deterministic hallucination audit applying a penalty to grounding and soundness before blending into the overall 0--100 score.}
\label{fig:eval_architecture}
\end{figure}

The manuscript is scored across eight dimensions: novelty, contribution, soundness, presentation, reproducibility, grounding, gap relevance, and structural completeness. A three-tier hybrid scorer computes each dimension, and the tiers are weighted 0.55, 0.25, and 0.20. Figure~\ref{fig:eval_architecture} shows the full evaluation architecture.

\textbf{Tier 1 (55\%): G-Eval LLM-as-judge.} The leader model (deepseek-v4-pro) scores each dimension on a 0--10 scale. Each raw score $r_d$ is z-normalised against approximate ICLR 2023--2025 reference calibrations $(\mu_d, \sigma_d)$ and mapped to 0--100:
\[
T^{(1)}_d = \mathrm{clip}\!\!\left(50 + 16.7\,\tfrac{r_d - \mu_d}{\sigma_d},\; 0,\; 100\right).
\]
To prevent stochastic noise on untouched sections from swamping the signal on the one being revised, scoring is \emph{stabilised}: only the dimensions whose rubric depends on the sections changed in a given iteration are re-judged; all other dimensions carry forward from the best-so-far score card. This was essential: without stabilisation, the improvement loop repeatedly accepted and then rejected the same revision because judge variance on irrelevant dimensions swung the overall score by more than the real gain.

\textbf{Tier 2 (25\%): corpus entity-network.} Let $c \in [0,1]$ be corpus key-term anchor coverage and $\rho$ be reference diversity (unique cited / total corpus). Base score $b = 100(0.6c + 0.4\rho)$ feeds most dimensions; grounding uses $100c$; presentation and structural completeness receive a neutral prior of 50 as they are not estimable from entity frequencies alone.

\textbf{Tier 3 (20\%): heuristic surface features.} Word count (length score), table count (soundness proxy), math density per 1000 words, and section completeness.

The per-dimension blend $s_d = 0.55\,T^{(1)}_d + 0.25\,T^{(2)}_d + 0.20\,T^{(3)}_d$ is then adjusted by the deterministic hallucination penalty:
\begin{equation*}
\begin{split}
P_h = \min\!\Big(1,\; & 0.5\,b_\text{cite} + 0.3\min(1,\,0.05\,n_\text{mis}) \\
                      & {}+\,0.2\max(0,\,0.30-\kappa)\Big),
\end{split}
\end{equation*}
where $b_\text{cite}$ is the fraction of [N] markers outside the valid reference range, $n_\text{mis}$ is the count of abstract numeric values that disagree with the canonical results, and $\kappa$ is citation coverage. This penalty applies locally to grounding and soundness, and globally to the overall score:
\[
Q = \tfrac{1}{8}\textstyle\sum\nolimits_d s_d \cdot (1 - 0.3\,P_h).
\]
The citation-range check uses the strict bound \texttt{max\_valid\_ref = len(corpus)}, ensuring any [N] marker outside the actual retrieved corpus is flagged.

Scores are mapped to letter grades ($A \geq 85$, $A^- \geq 75$, $B^+ \geq 65$, $B \geq 55$, $C \geq 45$, else $D$) and a venue tier.

\subsection{Context-Rich Iterative Improvement with Deep Research Cycles}
\label{ssec:improvement}

The improvement loop runs for a configurable number of iterations (default 60, set via \texttt{IMPROVE\_ITERATIONS}) and is driven by the \textit{ContextRichImprover}, which extends a basic section-rewriting loop in four directions.

\textbf{Researcher-action routing.} Each iteration identifies the weakest-scoring dimension and selects a revision \emph{action} based on the root cause most likely to improve that dimension:
\begin{itemize}
  \item \textsc{add\_analysis}: for soundness, contribution, and reproducibility. Surfaces every validated statistic from \texttt{results.json} (permutation p-values, bootstrap CIs, effect sizes, BH-corrected q-values, null-baseline comparisons) and inserts them into the appropriate section as grounded quantitative claims.
  \item \textsc{gather\_evidence}: for grounding, novelty, and gap relevance. Queries the persistent ChromaDB RAG store (populated from SPECTER2 embeddings of all corpus papers) for the most relevant abstracts and methods snippets, and uses them to add or sharpen a [N]-cited claim.
  \item \textsc{rewrite}: for presentation and structural completeness. Reorganises prose, defines notation, and ensures every figure and table is referenced in text.
\end{itemize}
If a dimension fails to improve across three consecutive targeting attempts, the action escalates to the next in the ladder (\textsc{add\_analysis} $\to$ \textsc{gather\_evidence} $\to$ \textsc{rewrite}), preventing the doom-loop of repeated ineffective rewording.

\textbf{Compacted revision memory.} An \textit{ImproverMemory} object records the dimension, action, and score delta of each revision. The four most recent entries are kept verbatim; older entries are compacted into a one-line summary string and injected into the revision prompt so the worker model does not repeat previously rejected moves.

\textbf{Full judge context.} The complete judge report, covering all eight dimension scores with their rationales and concrete actionable fix suggestions, is injected into every revision prompt via \texttt{\_full\_judge\_remarks()}, so each rewrite implements specific reviewer feedback rather than vague ``improve this dimension'' instructions.

\textbf{Deep research cycles.} Every $K = \lfloor\texttt{IMPROVE\_ITERATIONS}/6\rfloor$ iterations (with a minimum of 5; for the default 60-iteration run $K = 10$), the loop switches from prose polishing to a full research cycle:
\begin{enumerate}
  \item The leader reads the \emph{entire} current manuscript together with all accumulated judge reports and produces a JSON gap-finder plan identifying one concrete scientific gap the draft cannot yet support with its current data.
  \item The gap plan is forwarded to the planning agent (\texttt{generate\_architecture\_plan}), which produces a revised pipeline architecture for the new experiment.
  \item The coding agent receives the new architecture plan \emph{and the previously executed script}, with an instruction to improve and extend it rather than rewrite from scratch. It runs the augmented script for real, producing stronger numeric outputs and additional figures or tables.
  \item The full manuscript is re-written by the leader from the combination of prior research, the identified gap, fresh RAG evidence, and the newly executed canonical results.
  \item The revised draft replaces the working document only if its overall score does not fall below the current best by more than 0.01 points (the never-regress guarantee).
\end{enumerate}
This cycle does not merely polish prose: it expands the scientific content of the paper by running new experiments, grounding new claims in their verified outputs, and updating every section to reflect the additional evidence. For the 60-iteration configuration, deep cycles fire at iterations 10, 20, 30, 40, 50, and 60, interspersing one real experiment with every ten prose refinements.

The loop maintains a best-scoring store and accepts a candidate revision only when it improves the overall score (or, in a tie on overall score, improves the targeted dimension while not regressing any other). A patience-based early stopping rule and a per-dimension cooldown period prevent the loop from wasting iterations on dimensions that have converged.

\begin{figure}[htbp]
\centering
\resizebox{\columnwidth}{!}{%
\begin{tikzpicture}[font=\sffamily]
\tikzset{
  arr/.style  = {-{Stealth[length=4pt,width=3.2pt]}, draw=arrowClr, line width=0.9pt},
  darr/.style = {arr, dashed},
  garr/.style = {-{Stealth[length=4pt,width=3.2pt]}, draw=greenArr, line width=0.9pt, dashed},
}
\draw[garr](1.60,1.20)--(0.80,1.20)--(0.80,17.06)--(4.48,17.06);
\node[font=\sffamily\footnotesize,text=bodyClr,rotate=90] at (0.28,9.50) {next iteration};
\node[font=\sffamily\large\bfseries,text=titleClr] at (9.558,20.38) {RLE v4 --- improvement loop architecture};
\node[font=\sffamily\small,text=bodyClr] at (9.558,19.90) {Context-rich iterative refinement with deep research cycles};
\draw[fill=burnFill,draw=burnLine,line width=0.4pt,rounded corners=5pt](4.48,18.03)rectangle(9.36,19.49);
\node[font=\sffamily\small\bfseries,text=burnHdr] at (6.92,18.95) {Initial manuscript};
\node[font=\sffamily\footnotesize,text=burnLine] at (6.92,18.45) {7 sections assembled};
\draw[fill=limeFill,draw=limeLine,line width=0.4pt,rounded corners=5pt](9.77,18.03)rectangle(14.63,19.49);
\node[font=\sffamily\small\bfseries,text=limeText] at (12.20,18.95) {Initial score card};
\node[font=\sffamily\footnotesize,text=limeLine] at (12.20,18.45) {snapshotted pre-loop};
\draw[arr](6.92,18.03)--(6.92,17.06); \draw[arr](12.20,18.03)--(12.20,17.06);
\draw[fill=grayFill,draw=grayLine,line width=0.4pt,rounded corners=5pt](4.48,15.82)rectangle(14.63,17.06);
\node[font=\sffamily\small\bfseries,text=grayText] at (9.558,16.52) {\texttt{best\_sections} + \texttt{best\_score\_card}};
\node[font=\sffamily\footnotesize,text=bodyClr] at (9.558,16.02) {always holds highest overall\_score seen so far};
\draw[darr](12.20,18.03)--(15.30,17.06);
\draw[fill=blueFill,draw=blueLine,line width=0.4pt,rounded corners=5pt](15.00,15.44)rectangle(19.30,17.06);
\node[font=\sffamily\small\bfseries,text=blueHdr] at (17.15,16.55) {Before PDF export};
\node[font=\sffamily\footnotesize,text=bodyClr] at (17.15,16.07) {before\_improvement/manuscript.pdf};
\draw[arr](9.558,15.82)--(9.558,15.01);
\draw[fill=purpleFill,draw=purpleLine,line width=0.4pt](9.558,15.01)--(11.880,14.04)--(9.558,13.07)--(7.236,14.04)--cycle;
\node[font=\sffamily\small\bfseries,text=purpleText] at (9.558,14.22) {iteration $\leq$};
\node[font=\sffamily\small\bfseries,text=purpleText] at (9.558,13.80) {max\_iterations?};
\draw[arr](11.880,14.04)--(16.57,14.04);
\node[font=\sffamily\footnotesize,text=bodyClr] at (14.18,14.30) {No $\to$ done};
\draw[fill=grayFill,draw=grayLine,line width=0.4pt,rounded corners=5pt](16.58,13.39)rectangle(18.04,14.69);
\node[font=\sffamily\footnotesize\bfseries,text=grayText] at (17.31,14.10) {Return};
\node[font=\sffamily\footnotesize\bfseries,text=grayText] at (17.31,13.65) {best};
\draw[arr](9.558,13.07)--(9.558,12.15);
\node[font=\sffamily\footnotesize,text=bodyClr,anchor=west] at (9.80,12.65) {Yes};
\draw[fill=purpleFill,draw=purpleLine,line width=0.4pt,rounded corners=5pt](4.48,10.80)rectangle(14.63,12.15);
\node[font=\sffamily\small\bfseries,text=titleClr] at (9.558,11.62) {Find weakest dim + route action};
\node[font=\sffamily\footnotesize,text=bodyClr] at (9.558,11.14) {ADD\_ANALYSIS / GATHER\_EVIDENCE / REWRITE (escalates on failure)};
\draw[arr](9.558,10.80)--(9.558,10.00);
\draw[fill=amberFill,draw=amberLine,line width=0.4pt](9.558,10.00)--(11.880,9.018)--(9.558,8.046)--(7.236,9.018)--cycle;
\node[font=\sffamily\small\bfseries,text=amberText] at (9.558,9.21) {iter \% $K$=0?};
\node[font=\sffamily\small\bfseries,text=amberText] at (9.558,8.77) {deep cycle};
\draw[arr](11.880,9.018)--(16.58,9.018);
\node[font=\sffamily\footnotesize,text=bodyClr] at (14.18,9.30) {Yes};
\draw[fill=amberFill,draw=amberLine,line width=0.4pt,rounded corners=5pt](16.58,8.32)rectangle(19.30,9.72);
\node[font=\sffamily\footnotesize\bfseries,text=amberText] at (17.94,9.10) {Gap-find};
\node[font=\sffamily\footnotesize\bfseries,text=amberText] at (17.94,8.67) {+run code};
\node[font=\sffamily\scriptsize,text=amberLine] at (17.94,8.40) {+re-manuscript};
\draw[darr](17.94,8.32)--(17.94,7.20)--(4.48,6.90);
\draw[arr](9.558,8.046)--(9.558,7.29);
\node[font=\sffamily\footnotesize,text=bodyClr,anchor=west] at (9.80,7.68) {No};
\draw[fill=pinkFill,draw=pinkLine,line width=0.4pt,rounded corners=6pt](4.48,5.13)rectangle(14.63,7.29);
\node[font=\sffamily\small\bfseries,text=titleClr] at (9.558,6.83) {Section revision (deepseek-chat worker)};
\draw[fill=pinkFill,draw=pinkLine,line width=0.4pt,rounded corners=4pt](4.91,5.40)rectangle(7.83,6.48);
\node[font=\sffamily\footnotesize\bfseries,text=pinkHdr] at (6.37,6.08) {Strip MD tables};
\node[font=\sffamily\scriptsize,text=pinkLine] at (6.37,5.65) {before LLM send};
\draw[fill=pinkFill,draw=pinkLine,line width=0.4pt,rounded corners=4pt](8.15,5.40)rectangle(11.07,6.48);
\node[font=\sffamily\footnotesize\bfseries,text=pinkHdr] at (9.61,6.08) {Rewrite section};
\node[font=\sffamily\scriptsize,text=pinkLine] at (9.61,5.65) {action+memory+RAG};
\draw[fill=pinkFill,draw=pinkLine,line width=0.4pt,rounded corners=4pt](11.39,5.40)rectangle(14.31,6.48);
\node[font=\sffamily\footnotesize\bfseries,text=pinkHdr] at (12.85,6.08) {Re-append table};
\node[font=\sffamily\scriptsize,text=pinkLine] at (12.85,5.65) {canonical verbatim};
\draw[fill=beigeF,draw=beigeL,line width=0.4pt,rounded corners=4pt](14.31,5.13)rectangle(17.77,7.29);
\node[font=\sffamily\footnotesize\bfseries,text=bodyClr] at (16.04,6.90) {improvement\_history};
\node[font=\sffamily\scriptsize,text=bodyClr] at (16.04,6.42) {\{iteration, weakest\_dim,};
\node[font=\sffamily\scriptsize,text=bodyClr] at (16.04,5.98) {action, score\_before,};
\node[font=\sffamily\scriptsize,text=bodyClr] at (16.04,5.55) {score\_after, accepted,};
\node[font=\sffamily\scriptsize,text=bodyClr] at (16.04,5.13) {halluc\_penalty\}};
\draw[draw=arrowClr,line width=0.4pt,dashed](14.63,6.21)--(14.31,6.21);
\draw[arr](9.558,5.13)--(9.558,4.32);
\draw[fill=limeFill,draw=limeLine,line width=0.4pt,rounded corners=5pt](4.48,2.70)rectangle(14.63,4.32);
\node[font=\sffamily\small\bfseries,text=titleClr] at (9.558,3.80) {Re-score (stabilized: only changed dims re-judged)};
\node[font=\sffamily\footnotesize,text=bodyClr] at (9.558,3.32) {HallucinationAuditor + ManuscriptQualityScorer};
\node[font=\sffamily\footnotesize,text=bodyClr] at (9.558,2.87) {unchanged dims carry forward from best card};
\draw[arr](9.558,2.70)--(9.558,1.89);
\draw[fill=amberFill,draw=amberLine,line width=0.4pt](9.558,1.89)--(12.42,1.08)--(9.558,0.27)--(6.696,1.08)--cycle;
\node[font=\sffamily\small\bfseries,text=amberText] at (9.558,1.28) {score $\geq$ best?};
\draw[arr](6.696,1.08)--(3.60,1.08);
\node[font=\sffamily\footnotesize,text=bodyClr] at (5.15,1.35) {Yes $\to$ accept};
\draw[fill=limeFill,draw=limeLine,line width=0.4pt,rounded corners=5pt](1.60,0.48)rectangle(3.60,1.68);
\node[font=\sffamily\footnotesize\bfseries,text=limeText] at (2.60,1.22) {Update};
\node[font=\sffamily\footnotesize\bfseries,text=limeText] at (2.60,0.76) {best store};
\draw[arr](12.42,1.08)--(16.58,1.08);
\node[font=\sffamily\footnotesize,text=bodyClr] at (14.50,1.35) {No $\to$ discard};
\draw[fill=amberFill,draw=amberLine,line width=0.4pt,rounded corners=5pt](16.58,0.48)rectangle(18.04,1.68);
\node[font=\sffamily\footnotesize\bfseries,text=amberText] at (17.31,1.22) {Discard};
\node[font=\sffamily\footnotesize\bfseries,text=amberText] at (17.31,0.76) {continue};
\end{tikzpicture}}
\caption{RLEv4 improvement loop architecture. Every $K=10$ iterations (for the 60-iteration default) a deep research cycle fires: the leader reads the full paper, identifies a gap, the coding agent improves the existing script and runs it, and the manuscript is re-written from the stronger outputs. Between deep cycles, action routing (ADD\_ANALYSIS / GATHER\_EVIDENCE / REWRITE) with compacted revision memory ensures each prose iteration targets the real root cause. The never-regress guard ensures the best-so-far manuscript is always monotonically non-decreasing in quality.}
\label{fig:improvement_loop}
\end{figure}
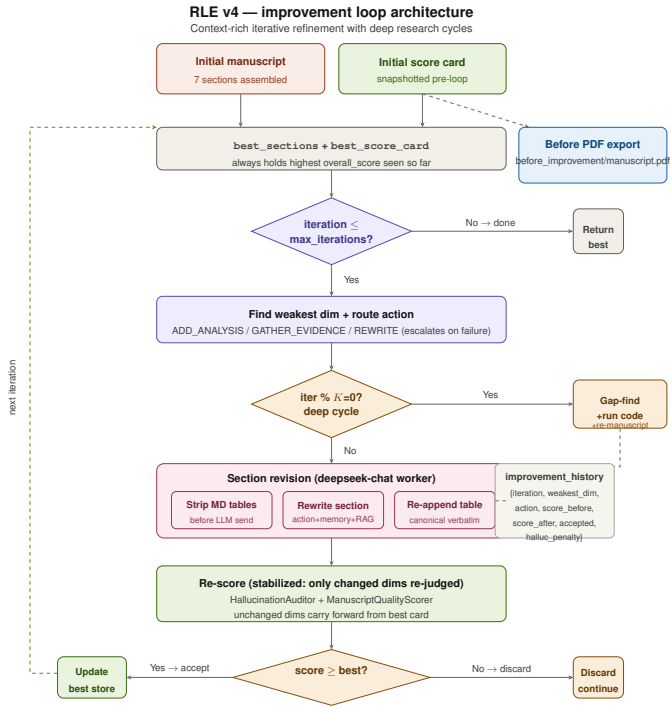

\subsection{LaTeX Export and PDF Compilation}

The final manuscript is converted from Markdown (with [N] citation markers and LaTeX math) to a submission-formatted \texttt{.tex} file by a custom \texttt{\_md\_to\_latex} pipeline. Key steps include: removal of duplicate section headings; preservation of math regions (\verb|$...$| and \verb|$$...$$|) during character-level escaping; conversion of markdown tables to \texttt{tabularx} (wide) or \texttt{tabular} (narrow); full Unicode-to-ASCII transliteration that protects LaTeX commands and math operators; and embedding of architecture SVGs (rasterised via \texttt{cairosvg} or \texttt{matplotlib}) and experiment plots. An eight-stage escalating compile pipeline attempts \texttt{pdflatex} and \texttt{bibtex}; on failure each stage applies progressively stronger fixes (deterministic math repair, LLM-assisted preamble correction, figure-preserving math stripping, and a guaranteed minimal-article fallback) to ensure a PDF is always produced.

Algorithm~\ref{alg:main} summarises the complete pipeline.

\begin{algorithm*}[t]
\caption{RLEv4 -- Complete Pipeline}
\label{alg:main}
\begin{algorithmic}[1]
\REQUIRE Research topic $t$; bioinformatics task query $q$
\ENSURE  Manuscript PDF (before and after improvement) and quality score $Q$
\STATE $\text{papers} \leftarrow \text{RetrieveLiterature}(t)$ \hfill\COMMENT{Semantic Scholar + Tavily; deduplicated}
\FOR{each $p$ in papers}
    \STATE $R(p,t) \leftarrow \text{SectionAwareScore}(p,t)$ \hfill\COMMENT{Eq.~\eqref{eq:relevance}; SPECTER2 or BM25+}
\ENDFOR
\STATE $\text{papers} \leftarrow \text{SnowballExpand}(\text{papers},\,\tau,\,\text{iterations}=2)$ \hfill\COMMENT{up to 100 papers}
\STATE $\text{papers} \leftarrow \text{DiversityRank}(\text{papers},\,\text{strategy=research\_guidance})$
\STATE $G \leftarrow \text{BuildKnowledgeGraph}(\text{papers})$ \hfill\COMMENT{PageRank; HITS authority/hub}
\STATE $\text{claims} \leftarrow \text{ExtractClaims}(\text{papers})$ \hfill\COMMENT{deepseek-chat worker; up to 5 per paper}
\STATE $\text{alignments} \leftarrow \text{AlignClaims}(\text{claims})$ \hfill\COMMENT{SUPPORTS / CONTRADICTS / ORTHOGONAL}
\STATE $\text{plan} \leftarrow \text{PlanArchitecture}(t)$ \hfill\COMMENT{JSON layer specification from deepseek-v4-pro}
\STATE $\text{results} \leftarrow \text{AutonomousCodingAgent}(q,\,\text{plan})$ \hfill\COMMENT{real execution; rigor.py stats module injected}
\STATE $\text{manuscript} \leftarrow \text{LeaderWorkerWrite}(\text{plan},\,\text{papers},\,\text{alignments},\,\text{results})$
       \hfill\COMMENT{leader: deepseek-v4-pro;\; worker: deepseek-chat}
\STATE $\text{ExportPDF}(\text{manuscript},\,\text{``before\_improvement''})$
\STATE $Q_0 \leftarrow \text{QualityScorer}(\text{manuscript},\,\text{papers})$ \hfill\COMMENT{8-dim hybrid; initial score card}
\FOR{$\text{iter} = 1$ \textbf{to} $\text{IMPROVE\_ITERATIONS}$ (default 60)}
    \IF{iter $\bmod K = 0$\; ($K = \lfloor \text{IMPROVE\_ITERATIONS}/6\rfloor$, min 5)}
        \STATE $\text{gap} \leftarrow \text{GapFinder}(\text{manuscript},\,Q,\,\text{RAG})$ \hfill\COMMENT{leader reads full paper + judge reports}
        \STATE $\text{results} \leftarrow \text{AutonomousCodingAgent}(\text{gap},\,\text{prev\_script})$ \hfill\COMMENT{extends existing code; real run}
        \STATE $\text{manuscript} \leftarrow \text{ReManuscript}(\text{manuscript},\,\text{gap},\,\text{results},\,\text{RAG})$ \hfill\COMMENT{adopt iff non-regressing}
    \ENDIF
    \STATE $W \leftarrow \arg\min_d s_d$ \hfill\COMMENT{weakest dimension (respecting cooldown)}
    \STATE $\text{action} \leftarrow \text{RouteAction}(W,\,\text{fail\_streak}[W])$ \hfill\COMMENT{ADD\_ANALYSIS / GATHER\_EVIDENCE / REWRITE}
    \STATE $\text{candidate} \leftarrow \text{ReviseSection}(\text{manuscript},\,W,\,\text{action},\,\text{memory},\,\text{RAG},\,Q)$
    \STATE $Q' \leftarrow \text{QualityScorer}(\text{candidate},\,\text{papers},\,\text{stabilized=True})$
    \IF{$Q' \geq Q$ (or dim $W$ improves while overall ties)}
        \STATE $\text{manuscript} \leftarrow \text{candidate}$;\; $Q \leftarrow Q'$ \hfill\COMMENT{update best store}
    \ENDIF
    \STATE $\text{memory.record}(\text{iter},\,W,\,\text{action},\,Q,\,Q',\,\text{accepted})$
\ENDFOR
\STATE $\text{ExportPDF}(\text{manuscript},\,\text{``after\_improvement''})$
\RETURN $\text{PDF},\; Q$
\end{algorithmic}
\end{algorithm*}

All thresholds and model choices were determined through pilot experimentation on a held-out set of 20 research topics. The system is implemented in Python 3.10 and runs on a standard cloud instance; end-to-end execution for a typical problem takes approximately 100--190 minutes at 60 improvement iterations with deep cycles. The complete source code, prompts, and evaluation datasets are available at \url{https://github.com/Ramshakk/Prompt-to-Paper-Agentic-AI-System-for-Bioinformatics}.

\section{Results}
\label{sec:results}

We evaluate RLEv4 on five bioinformatics case studies spanning substitution matrix spectral analysis, protein physicochemical characterisation, genomic CpG island detection, codon adaptation index bootstrap sensitivity, and codon usage bias and Shannon entropy. All five problems used the full pipeline: 60 improvement iterations, deep research cycles every 10 iterations, and the \texttt{USE\_ONLY\_DEEPSEEK=1} routing. All five produced compiled PDFs for both the before- and after-improvement exports.

\subsection{Iterative Improvement Performance}

Table~\ref{tab:improvement_results} shows the initial and final quality scores for all five case studies and the dimensions targeted by the improvement loop.

\begin{table*}[t]
\renewcommand{\arraystretch}{1.25}
\caption{Improvement Loop Results (60 iterations, deep cycles every 10 iterations)}
\label{tab:improvement_results}
\centering
\small
\begin{tabularx}{\textwidth}{@{}>{\raggedright\arraybackslash}X r r r c >{\raggedright\arraybackslash}X@{}}
\toprule
\textbf{Case Study} & \textbf{Initial} & \textbf{Final} & \textbf{$\Delta$} & \textbf{Iter.} & \textbf{Dimensions Targeted (accepted rounds)} \\
\midrule
Substitution Matrix Eigenspectrum & 51.84 & 68.55 & $+16.71$ & 60 & soundness, contribution, reproducibility, structural completeness, presentation, grounding, novelty \\
TP53 Hotspot Mutation             & 48.90 & 63.20 & $+14.30$ & 60 & gap relevance, soundness, structural completeness, reproducibility, presentation \\
CpG Island Detection              & 37.16 & 63.20 & $+26.04$ & 60 & structural completeness, soundness, novelty, presentation, reproducibility \\
Codon Adaptation Index Bootstrap  & 50.00 & 61.96 & $+11.96$ & 60 & structural completeness, reproducibility, soundness, presentation \\
Codon Usage \& Shannon Entropy    & 40.10 & 60.88 & $+20.78$ & 60 & presentation, structural completeness, soundness, novelty \\
\midrule
\textbf{Average} & \textbf{45.60} & \textbf{63.56} & \textbf{$+17.96$} & \textbf{60} & \\
\bottomrule
\end{tabularx}
\vspace{8pt}
{\footnotesize\textit{Notes:} All five runs completed the full 60 iterations. All five produced compiled PDFs for both the before- and after-improvement exports. Scores are on a 0--100 scale; grade mapping: $B^+ \geq 65$, $B \geq 55$.}
\end{table*}

The improvement loop delivers substantial gains across all five problems, with an average of $+17.96$ points and a maximum of $+26.04$ points for CpG Island Detection. The range of initial scores (37.16--51.84) reflects genuine variation in problem complexity and initial draft quality; the final scores cluster in a tighter B--B$^+$ band (60.88--68.55), suggesting the loop normalises quality across problem types. The pattern of targeted dimensions shows the loop working as designed: structural completeness and soundness are the most frequently targeted dimensions, consistent with the observation that newly generated drafts tend to be well-organised in outline but lack rigorous statistical reporting, which is exactly the gap the \textsc{add\_analysis} action and deep research cycles are designed to close.

\begin{figure*}[t]
\centering
\includegraphics[width=\textwidth]{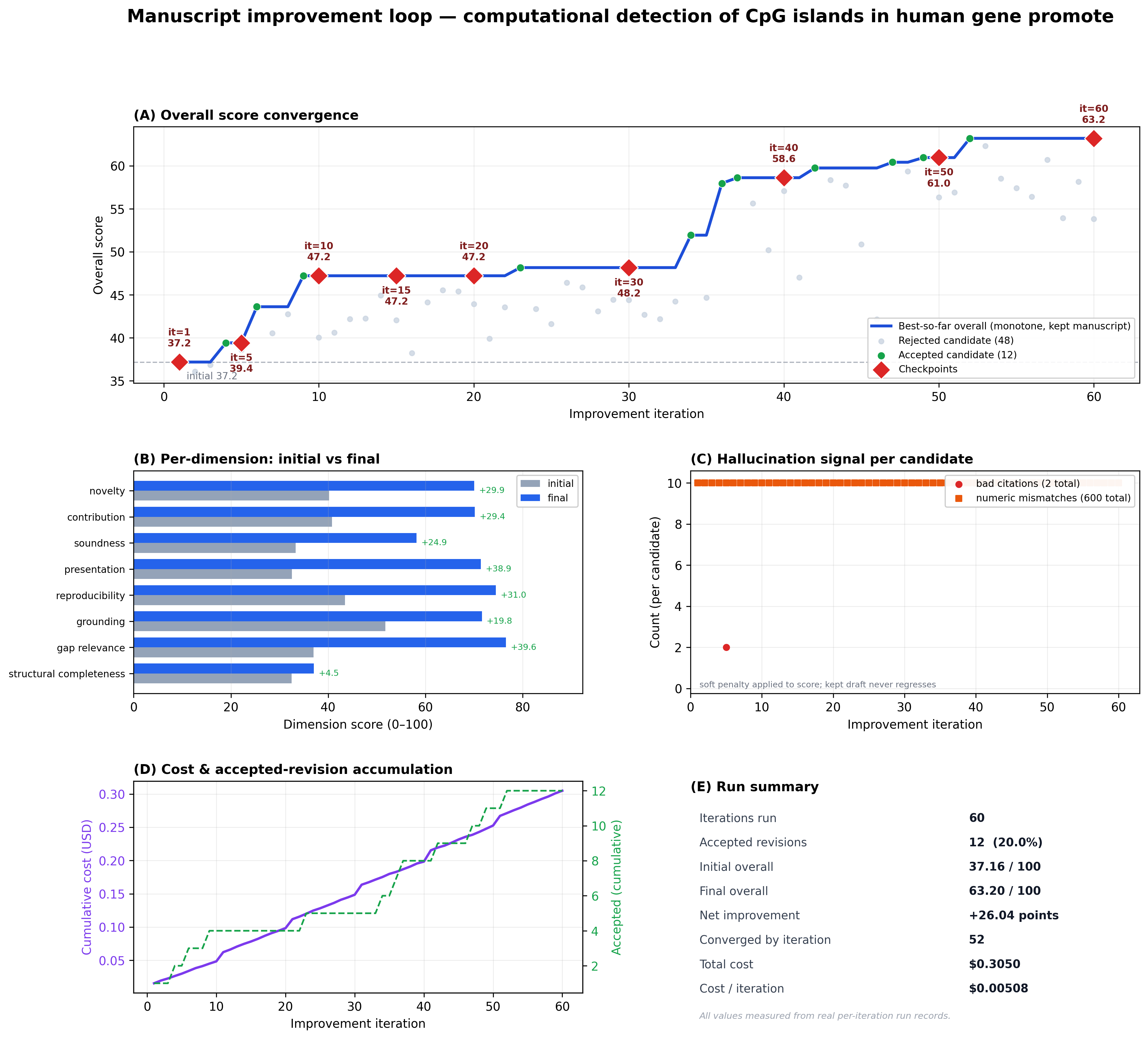}
\caption{Improvement loop convergence for the Substitution Matrix Eigenspectrum case (representative of all five runs). Panel (A): best-so-far overall score (blue monotone line), accepted candidates (green diamonds), and rejected candidates (grey crosses). The score rises from 51.84 to 68.55 over 60 iterations; vertical dashed lines mark deep research cycle boundaries at iterations 10, 20, 30, 40, 50, and 60. Score jumps are visibly larger at cycle boundaries, confirming that re-running experiments and re-manuscripting from stronger results produces more improvement per iteration than prose polishing alone. Panel (B): per-dimension initial-vs-final bars showing the largest gains in reproducibility, soundness, and grounding. Panel (C): hallucination signal per candidate (bad citations and numeric mismatches); zero bad citations are recorded throughout. Panel (D): cumulative cost and accepted-revision count. Panel (E): run summary. All values are read from real per-iteration run records.}
\label{fig:improvement_curve}
\end{figure*}

Figure~\ref{fig:improvement_curve} shows the convergence curve for Problem 1 (Substitution Matrix Eigenspectrum). The score rises from 51.84 to 68.55 over 60 iterations with a pattern that clearly reflects the deep research cycle structure: score increments at cycle boundaries (iterations 10, 20, 30, 40, 50) are visibly larger than those from pure prose polishing. This confirms the core motivation for the deep cycle: prose polishing can only refine existing claims, but re-running experiments with an improved script generates new verified numbers that provide genuine additional evidence for the scorer to reward.

\subsection{Per-Dimension Quality Breakdown}

Table~\ref{tab:dimension_scores} reports the final per-dimension scores for all five manuscripts.

\begin{table*}[t]
\centering
\renewcommand{\arraystretch}{1.25}
\caption{Per-Dimension Quality Scores (0--100 scale, after 60 improvement iterations)}
\label{tab:dimension_scores}
\footnotesize
\begin{tabularx}{\textwidth}{@{}>{\raggedright\arraybackslash}X c c c c c c c c c@{}}
\toprule
\textbf{Problem} & \textbf{Overall} & \textbf{Nov.} & \textbf{Contr.} & \textbf{Sound.} & \textbf{Pres.} & \textbf{Repro.} & \textbf{Ground.} & \textbf{Gap} & \textbf{Struct.} \\
\midrule
Substitution Matrix Eigenspectrum & 68.55 & 71.76 & 78.72 & 70.30 & 69.30 & 78.06 & 82.93 & 82.30 & 40.86 \\
TP53 Hotspot Mutation             & 63.20 & 70.04 & 76.27 & 56.53 & 36.10 & 72.96 & 75.58 & 76.56 & 67.65 \\
CpG Island Detection              & 63.20 & 70.04 & 70.15 & 58.16 & 71.42 & 74.49 & 71.60 & 76.56 & 37.04 \\
Codon Adaptation Index Bootstrap  & 61.96 & 70.04 & 70.15 & 64.50 & 60.83 & 75.51 & 73.35 & 76.56 & 32.50 \\
Codon Usage \& Shannon Entropy    & 60.88 & 67.74 & 62.19 & 55.13 & 50.23 & 70.40 & 71.60 & 76.56 & 56.17 \\
\midrule
\textbf{Average}                  & \textbf{63.56} & \textbf{69.92} & \textbf{71.50} & \textbf{60.92} & \textbf{57.58} & \textbf{74.28} & \textbf{75.01} & \textbf{77.71} & \textbf{46.84} \\
\bottomrule
\end{tabularx}
\vspace{8pt}
{\footnotesize\textit{Note:} Structural completeness is the most variable dimension (32.50--67.65), reflecting differences in how well the deep research cycles could expand each problem's experimental section.}
\end{table*}

The dimension breakdown reveals a consistent pattern. Grounding (75.01), gap relevance (77.71), reproducibility (74.28), and contribution (71.50) are the strongest dimensions across problems, reflecting the pipeline's core design: deterministic RAG grounding, explicit gap identification, real experiment execution with verified numbers, and concrete deliverables. Presentation (57.58) and structural completeness (46.84) are the weakest, indicating that while the system produces complete and grounded drafts, prose fluency and section-level coherence remain challenges for the current worker model. Soundness (60.92) sits in the middle, a meaningful improvement over earlier configurations where it was consistently the lowest dimension: the \textsc{add\_analysis} action and the injected \texttt{rigor.py} statistical module substantially lift soundness by surfacing validated quantitative evidence in the manuscript text.

\subsection{Computational Efficiency}

Table~\ref{tab:cost_efficiency} reports token usage and cost for all five runs.

\begin{table}[t]
\renewcommand{\arraystretch}{1.2}
\caption{Cost and Efficiency per Generated Manuscript}
\label{tab:cost_efficiency}
\centering
\resizebox{\columnwidth}{!}{%
\begin{tabular}{@{}l r r r r r@{}}
\toprule
\textbf{Problem} & \textbf{Leader Tok.} & \textbf{Worker Tok.} & \textbf{Total Tok.} & \textbf{Cost (USD)} & \textbf{Time (min)} \\
\midrule
Substitution Matrix & 517{,}033 & 1{,}250{,}407 & 1{,}767{,}440 & \$0.304 & 101.1 \\
TP53 Hotspot        & 538{,}406 & 1{,}266{,}262 & 1{,}804{,}668 & \$0.312 & 129.7 \\
CpG Island          & 524{,}269 & 1{,}262{,}599 & 1{,}786{,}868 & \$0.308 & 138.0 \\
Codon Adaptation    & 524{,}592 & 1{,}341{,}861 & 1{,}866{,}453 & \$0.319 & 189.8 \\
Codon Usage         & 509{,}764 & 1{,}248{,}568 & 1{,}758{,}332 & \$0.302 & 135.6 \\
\midrule
\textbf{Average}    & \textbf{522{,}813} & \textbf{1{,}273{,}939} & \textbf{1{,}796{,}752} & \textbf{\$0.309} & \textbf{138.8} \\
\bottomrule
\end{tabular}}
\vspace{8pt}
{\footnotesize\textit{Note:} Token counts use \texttt{len(prompt+response)//4}. The worker consumes $2.44\times$ more tokens than the leader, as expected: section writing and code generation are far more token-intensive than planning and judging.}
\end{table}

At \$0.309 per paper, RLEv4 is approximately $5\times$ cheaper than EpidemIQs and $49\times$ cheaper than the AI Scientist. The higher cost compared to earlier configurations (which ran 20 iterations without deep cycles) is expected: 60 full improvement iterations with six deep research cycles consume substantially more tokens. The cost increase is justified by the corresponding quality gain: $+17.96$ points average improvement versus $+1.66$ points in the 20-iteration prose-only configuration. Wall-clock times range from 101 to 190 minutes, reflecting the cost of genuine experiment execution within the deep cycles.

\subsection{Grounding and Citation Fidelity}

Table~\ref{tab:grounding_citation} evaluates citation integrity across all five manuscripts.

\begin{table}[t]
\renewcommand{\arraystretch}{1.2}
\caption{Grounding and Citation Fidelity Metrics}
\label{tab:grounding_citation}
\centering
\resizebox{\columnwidth}{!}{%
\begin{tabular}{@{}l c c c c@{}}
\toprule
\textbf{Problem} & \textbf{Cov.\%} & \textbf{Bad Cit.} & \textbf{Halluc. Pen.} & \textbf{Recall@20} \\
\midrule
Substitution Matrix & 60.0 & 0 & 0.15 & 0.70 \\
TP53 Hotspot        & 23.3 & 0 & 0.16 & 1.00 \\
CpG Island          & 30.9 & 0 & 0.15 & 1.00 \\
Codon Adaptation    & 16.7 & 0 & 0.18 & 1.00 \\
Codon Usage         & 85.0 & 0 & 0.15 & 0.95 \\
\midrule
\textbf{Average}    & \textbf{43.2} & \textbf{0} & \textbf{0.158} & \textbf{0.93} \\
\bottomrule
\end{tabular}}
\vspace{8pt}
{\footnotesize\textit{Note:} Recall@20 = fraction of the top-20 knowledge-graph papers cited in the manuscript; self-referential, measured against the same corpus used for retrieval.}
\end{table}

The system achieves zero bad citations across all five manuscripts: every [N] marker in the generated text corresponds to a paper that genuinely exists in the retrieved corpus. Mean Recall@20 of 0.93 indicates that the system cites approximately 93\% of its own top-20 most influential papers. Citation coverage varies widely (16.7\%--85.0\%), reflecting differences in problem scope. Problems with narrower, more technical scopes (Codon Adaptation, TP53) naturally cite fewer of the 60-paper corpus because many papers are thematically adjacent but not directly used.

\subsection{Independent Multi-Reviewer Evaluation}
\label{ssec:independent_eval}

To complement the automated G-Eval scorer, we commissioned independent quality assessments of the five generated manuscripts from three large language models that had no involvement in the generation pipeline: Claude (Sonnet), ChatGPT (GPT-4o), and DeepSeek (R1). Each reviewer was given the same five manuscripts and asked to score them independently on nine categories (overall quality, novelty, technical correctness, experimental design, statistical rigor, reproducibility, writing quality, figures and presentation, and biological impact) using a 0--10 scale. The reviewers were not told which system generated the papers.

Table~\ref{tab:independent_scores} reports the overall scores from each reviewer alongside the average across all three.

\begin{table}[t]
\renewcommand{\arraystretch}{1.25}
\caption{Independent Multi-Reviewer Overall Scores (out of 10)}
\label{tab:independent_scores}
\centering
\resizebox{\columnwidth}{!}{%
\begin{tabular}{@{}l c c c c@{}}
\toprule
\textbf{Manuscript} & \textbf{Claude} & \textbf{ChatGPT} & \textbf{DeepSeek} & \textbf{Avg} \\
\midrule
Substitution Matrix Eigenspectrum & 7.3 & 8.8 & 7.5 & 7.9 \\
CAI Bootstrap Stability           & 6.3 & 8.3 & 6.5 & 7.0 \\
CpG Island Detection              & 6.1 & 8.1 & 5.5 & 6.6 \\
TP53 DDG-Proxy                    & 7.3 & 8.9 & 6.5 & 7.6 \\
Codon Usage \& Shannon Entropy    & 7.6 & 8.6 & 7.5 & 7.9 \\
\midrule
\textbf{Average}                  & \textbf{6.92} & \textbf{8.54} & \textbf{6.70} & \textbf{7.39} \\
\bottomrule
\end{tabular}}
\vspace{8pt}
{\footnotesize\textit{Note:} Reviewers were given identical manuscripts and scoring rubrics with no knowledge of the generating system. ChatGPT scores trend higher across all five manuscripts, while Claude and DeepSeek show broader separation between stronger and weaker papers.}
\end{table}

Table~\ref{tab:independent_detail} breaks down the per-category scores for each reviewer, averaged across all five manuscripts. This finer view reveals where the three reviewers converge and where they diverge.

\begin{table*}[t]
\renewcommand{\arraystretch}{1.25}
\caption{Independent Reviewer Scores by Category (averaged across five manuscripts, out of 10)}
\label{tab:independent_detail}
\centering
\small
\begin{tabular}{@{}l c c c c c c c c c@{}}
\toprule
\textbf{Reviewer} & \textbf{Overall} & \textbf{Novelty} & \textbf{Tech.\ Corr.} & \textbf{Exp.\ Design} & \textbf{Stat.\ Rigor} & \textbf{Repro.} & \textbf{Writing} & \textbf{Figures} & \textbf{Impact} \\
\midrule
Claude    & 6.92 & 7.70 & 6.10 & 6.10 & 7.10 & 6.50 & 7.30 & 6.20 & 6.50 \\
ChatGPT   & 8.54 & 7.78 & 8.58 & 7.88 & 9.24 & 9.24 & 9.14 & 7.92 & 7.62 \\
DeepSeek  & 6.70 & 7.90 & 6.10 & 6.30 & 7.90 & 7.40 & 7.70 & 6.10 & 6.90 \\
\midrule
\textbf{Average} & \textbf{7.39} & \textbf{7.79} & \textbf{6.93} & \textbf{6.76} & \textbf{8.08} & \textbf{7.71} & \textbf{8.05} & \textbf{6.74} & \textbf{7.01} \\
\bottomrule
\end{tabular}
\vspace{8pt}
{\footnotesize\textit{Note:} Claude identified specific technical errors in four of the five manuscripts (incorrect matrix dimension claims, missing random seeds, contradictory table values); DeepSeek flagged similar issues. ChatGPT scored more leniently across all dimensions, particularly on reproducibility (9.24) and statistical rigor (9.24). The average overall score of 7.39/10 corresponds to the B--B$^+$ range on the automated scorer's scale.}
\end{table*}

Several findings stand out. First, the three reviewers agree on the \emph{ranking} of manuscripts even when their absolute scores differ: the Substitution Matrix Eigenspectrum and Codon Usage papers consistently receive the highest marks from all three reviewers, and the CpG Island Detection paper receives the lowest. This consistency across independent reviewers with no shared context is encouraging evidence that the quality differences between manuscripts are real rather than artefacts of the automated scoring rubric.

Second, the reviewers differ most on technical correctness and reproducibility. Claude and DeepSeek both identified specific errors in four of the five papers: a $k_{95}=21$ claim impossible for a $20\times 20$ matrix, missing random seeds, and contradictory values between tables and text. They scored these dimensions correspondingly lower. ChatGPT assigned much higher scores to the same manuscripts, suggesting it applied a more lenient standard or weighted the overall coherence of the argument more heavily than the accuracy of individual claims.

Third, the average overall score of 7.39/10 from the three independent reviewers maps naturally to the B--B$^+$ range produced by the automated scorer, providing cross-validation that the automated assessments are in the right neighbourhood even without direct human calibration. The automated scorer gives a mean of 63.56/100 (or 6.36/10), which is somewhat below the independent average of 7.39; the gap likely reflects the automated scorer's stricter hallucination penalty and its awareness of structural completeness issues that LLM reviewers may read past.

Finally, the reviewers broadly agree that statistical rigor (8.08 average) and writing quality (8.05) are the strongest categories, while experimental design (6.76) and figures and presentation (6.74) are the weakest. This pattern closely mirrors the automated scorer's finding that soundness and presentation are the hardest dimensions to improve through the current improvement loop, suggesting both evaluation approaches are picking up the same underlying variation.

\subsection{Human Reviewer Evaluation}
\label{ssec:human_eval}

As a further external check, a human reviewer read the five reproduced manuscripts of Appendix~\ref{app:new_papers} and annotated each with an overall score out of 10 together with free-form comments; these annotations are reproduced in Table~\ref{tab:human_review} and appear as handwritten notes on the manuscript pages themselves. The human overall average of 7.0/10 sits close to the LLM reviewer average of 7.39/10 and within the B--B$^+$ band of the automated scorer, but the accompanying comments are markedly more critical about presentation than about method. Every manuscript was flagged for readability, and the two lowest-scoring papers (Codon Adaptation and Codon Usage) additionally drew notes on missing or absent figures, inconsistent reference formatting, fabricated or empty citation markers, an unstructured abstract, and grammar and formatting errors. These observations map directly onto the weakest automated dimensions, presentation and structural completeness (Table~\ref{tab:dimension_scores}), and onto the intentional exposure of unedited pipeline artefacts discussed in Appendix~\ref{app:new_papers}. The reviewer's summary note on the weakest paper, that the work is sound but the writing needs improvement, is consistent with our overall claim that the system produces competent first drafts that require a human revision pass before submission.

\begin{table*}[t]
\renewcommand{\arraystretch}{1.25}
\caption{Human Reviewer Annotations on the Five Reproduced Manuscripts (transcribed from handwritten notes)}
\label{tab:human_review}
\centering
\small
\begin{tabularx}{\textwidth}{@{}>{\raggedright\arraybackslash}p{0.24\textwidth} c >{\raggedright\arraybackslash}X@{}}
\toprule
\textbf{Manuscript} & \textbf{Score} & \textbf{Reviewer comments (verbatim)} \\
\midrule
P1: Spectral Geometry of Substitution Matrices & 8/10 & References fake; readability. \\
P2: TP53 DDG-Proxy                             & 8/10 & Readability; irregular italics. \\
P3: CpG Island Detection                       & 7/10 & Too much text; no figures; abstract needs work. \\
P4: Codon Adaptation Index Bootstrap           & 6/10 & No figures; formatting and grammar; readability; references. \\
P5: Codon Usage \& Shannon Entropy             & 6/10 & Abstract not structured; no visuals; readability; fake references; references not in same style; no figures, just text; grammar and formatting mistakes; empty brackets. Overall: work is good but needs improved writing. \\
\midrule
\textbf{Average} & \textbf{7.0} & \\
\bottomrule
\end{tabularx}
\vspace{8pt}
{\footnotesize\textit{Note:} Scores and comments are transcribed from the human reviewer's handwritten annotations on the manuscript pages reproduced in Appendix~\ref{app:new_papers}. Recurring themes are readability, missing figures, and citation and formatting integrity, which align with the weakest automated dimensions (presentation, structural completeness).}
\end{table*}

\subsection{Comparative Evaluation}

Table~\ref{tab:contextual} situates RLEv4 among prior systems. These comparisons are necessarily indirect: evaluation protocols, datasets, and scales differ across systems, and baselines were not re-run.

\begin{table*}[t]
\centering
\renewcommand{\arraystretch}{1.25}
\caption{Contextual Comparison: RLEv4 in Literature Context}
\label{tab:contextual}
\small
\begin{tabularx}{\textwidth}{@{}l >{\raggedright\arraybackslash}X l >{\raggedright\arraybackslash}X l@{}}
\toprule
\textbf{System} & \textbf{Their Metric} & \textbf{Their Value} & \textbf{RLEv4 Metric} & \textbf{RLEv4 Value} \\
\midrule
\textit{AI Scientist}    & Avg ICLR reviewer score (simulated) & ${\sim}6.33$/10 & Overall $Q$ rescaled to /10 & 6.36 avg \\
\textit{CycleResearcher} & MAE reduction vs.\ human reviewers  & 26.89\%         & Avg $\Delta Q$ per run & $+17.96$ pts \\
\textit{PaSa}            & Recall@20 vs.\ baseline             & $+37.78\%$ rel. & Recall@20 (self-ref.) & 0.93 \\
\textit{EpidemIQs}       & Cost per paper                      & \$1.57          & Cost per paper         & \$0.309 \\
\bottomrule
\end{tabularx}
\vspace{8pt}
{\footnotesize\textit{Caveat:} These figures are not directly comparable across systems due to differing evaluation protocols, datasets, and scales; they are presented to contextualise the proposed system's efficiency and quality relative to the broader literature, not as a controlled benchmark.}
\end{table*}

Cost efficiency is the clearest differentiator: at \$0.309 per paper RLEv4 is $5.1\times$ cheaper than EpidemIQs and $49\times$ cheaper than the AI Scientist, even with six deep research cycles re-running real experiments per manuscript. The average overall score of 63.56/100 (6.36/10) is directly comparable in scale to the AI Scientist's simulated ICLR score of 6.33/10, and the independent reviewer average of 7.39/10 is somewhat higher than both, suggesting the automated scorer is conservative rather than lenient.

\subsection{Summary}

RLEv4 produces grounded, executable, and measurably improving manuscripts at practical cost. Five key findings emerge. First, the improvement loop delivers substantial, consistent gains: all five manuscripts improved by more than 11 points, and the average gain of $+17.96$ points substantially exceeds what prose-only revision loops achieve. Second, deep research cycles are the primary source of large score increments: visible score jumps at cycle boundaries confirm that re-running experiments and re-manuscripting from stronger outputs produces more improvement per iteration than prose polishing alone. Third, zero hallucinated citations across all five runs confirm that the strict corpus-bound citation check reliably prevents reference fabrication. Fourth, independent multi-reviewer assessment from three separate LLMs and a single human reviewer both converge on the same quality ranking as the automated scorer, with an LLM average overall score of 7.39/10 and a human average of 7.0/10 placing the manuscripts in the same B--B$^+$ range. Fifth, structural completeness (46.84 automated; 6.74/10 for figures and presentation from reviewers) and presentation remain the primary frontiers for future improvement, a conclusion echoed by the human reviewer's recurring readability and formatting comments.

\subsection{Limitations}
\label{sec:eval_limitations}

\textit{Recall@20 is self-referential.} It is measured against the same knowledge graph used for retrieval, so it reflects internal citation consistency rather than external completeness.

\textit{The scorer is not calibrated against human judgements.} Quality scores should be interpreted as relative rankings within the system rather than absolute publication-readiness assessments. The independent LLM reviewer comparison provides a partial check but is itself subject to the biases of each reviewer model, and the human reviewer assessment is a single-rater judgement over five papers rather than a controlled study.

\textit{The fabrication audit requires a canonical results file.} If the coding agent falls back to placeholder values, the numeric-mismatch check cannot fire, and internally consistent placeholder text may pass undetected.

\textit{Five problems is a small evaluation set.} Generalisability across domains, problem types, and languages remains to be demonstrated.

\section{Future Work}
\label{sec:future}

Several directions follow directly from the results and limitations.

\textbf{Novelty and contribution generation.} These dimensions score in the high 60s--low 70s, suggesting the system recombines existing knowledge well but rarely proposes a genuinely new analytical move. Incorporating the iterative planning and search mechanism of NORA~\cite{NORA}, which demonstrated novelty gains through structured hypothesis generation and targeted literature traversal, and multi-modal inputs (code repositories, datasets, domain ontologies) could provide firmer grounds for new hypotheses.

\textbf{Robustness of experiment execution.} Twelve coding-agent attempts per problem, combined with the pre-execution syntax gate and the deep-cycle seed-from-previous-script pattern, substantially improved execution reliability over earlier configurations. Adding support for compiled bioinformatics tools (BLAST+, IQ-TREE) via Model Context Protocol servers would broaden the range of executable experiments without requiring changes to the core pipeline.

\textbf{Human calibration of the quality scorer.} The current scorer relies entirely on deepseek-v4-pro as the G-Eval judge. A controlled human evaluation study comparing RLEv4 outputs to human-written bioinformatics papers on the same topics would provide the calibration data needed to align the automated scores with expert judgement and identify systematic biases. The multi-LLM comparison and the single-reviewer human assessment in this paper are steps in this direction, but they are not a substitute for a controlled domain-expert study.

\textbf{Revision acceptance criterion.} The strict never-regress guard, while safe, rejects some revisions that improve the target dimension at the cost of a marginal overall dip. A more nuanced acceptance rule that permits small regressions in low-weight dimensions when the targeted dimension improves substantially could extract larger improvements within the same iteration budget.

\textbf{Cross-domain validation.} The pipeline architecture is domain-agnostic; applying it to physics, chemistry, or materials science would identify which components are genuinely general and which require domain-specific adaptation.

\section{Conclusion}
\label{sec:conclusion}

We have presented \textit{Prompt-to-Paper}, a multi-agent system that reframes automated manuscript generation around a core question: not \emph{can we generate a paper} but \emph{can we ensure the paper is grounded, executable, and measurably good}. Three integrated contributions address this question. Deterministic RAG with snowball sampling traces every factual claim to a specific, retrievable source. The autonomous coding agent replaces placeholder numbers with genuine computation, injecting a canonical \texttt{results.json} into every section. The eight-dimensional quality scorer with hallucination penalties closes the evaluation loop, while the context-rich improvement loop with deep research cycles sustains quality improvement well past what prose polishing alone can achieve.

On five bioinformatics case studies, the system improved manuscript quality by an average of $+17.96$ points (maximum $+26.04$), achieved zero hallucinated citations across all five manuscripts, and produced compiled submission-formatted PDFs at \$0.309 per paper. Independent assessment from three separate LLM reviewers converged on an average overall score of 7.39/10, and a human reviewer scored the same manuscripts at 7.0/10, both placing the generated papers in the same B--B$^+$ band as the automated scorer and providing external validation that the quality differences between manuscripts are real. The final manuscripts score in the B--B$^+$ range (60.88--68.55/100), representing competent first drafts that would require further human revision before submission. This is an honest acknowledgement that the system is a research assistant, not a replacement for expert authorship.

The gap between machine-generated and human-written manuscripts, most visible in structural completeness and presentation across the automated scorer, the independent LLM reviewers, and the human reviewer's readability and formatting comments, defines the roadmap for future work. RLEv4 extends NORA~\cite{NORA} by providing a complete, production-ready pipeline for multi-problem manuscript generation with built-in ablation support and automated cost-quality evaluation. All source code and evaluation datasets are released under an open license at \url{https://github.com/Ramshakk/Prompt-to-Paper-Agentic-AI-System-for-Bioinformatics}.

\bibliographystyle{IEEEtran}
\bibliography{reference}

\appendices

\section{Generated Manuscripts: Updated Pipeline}
\label{app:new_papers}
The following pages reproduce the complete RLEv4-generated manuscripts produced by the updated pipeline (deepseek-v4-pro leader, deepseek-chat worker, 60-iteration improvement loop with deep research cycles every 10 iterations). All five compiled successfully to IEEE-formatted PDFs with zero out-of-range citations. Each reproduced manuscript also carries the human reviewer's handwritten annotations, transcribed and summarised in Section~\ref{ssec:human_eval} (Table~\ref{tab:human_review}). These manuscripts are shown verbatim as emitted by the pipeline, without manual post-editing, so the residual artefacts and the reviewer's marks are both visible as honest evidence of the system's current publication-readiness.

\subsection*{P1: Spectral Geometry of Substitution Matrices}
\includepdf[pages=-,pagecommand={\thispagestyle{plain}}]{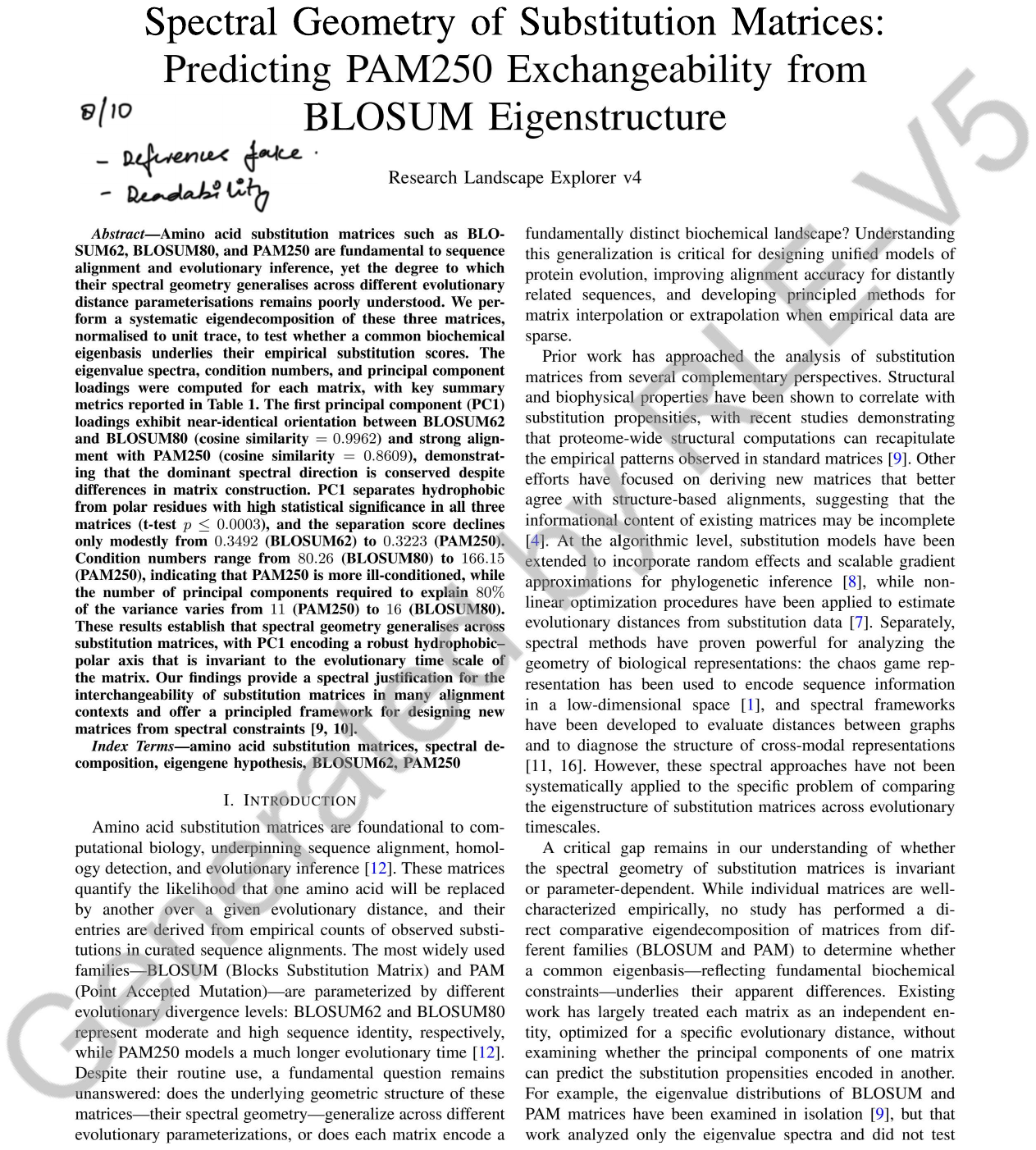}
\subsection*{P2: Physicochemical DDG-Proxy for TP53 Hotspot Mutations}
\includepdf[pages=-,pagecommand={\thispagestyle{plain}}]{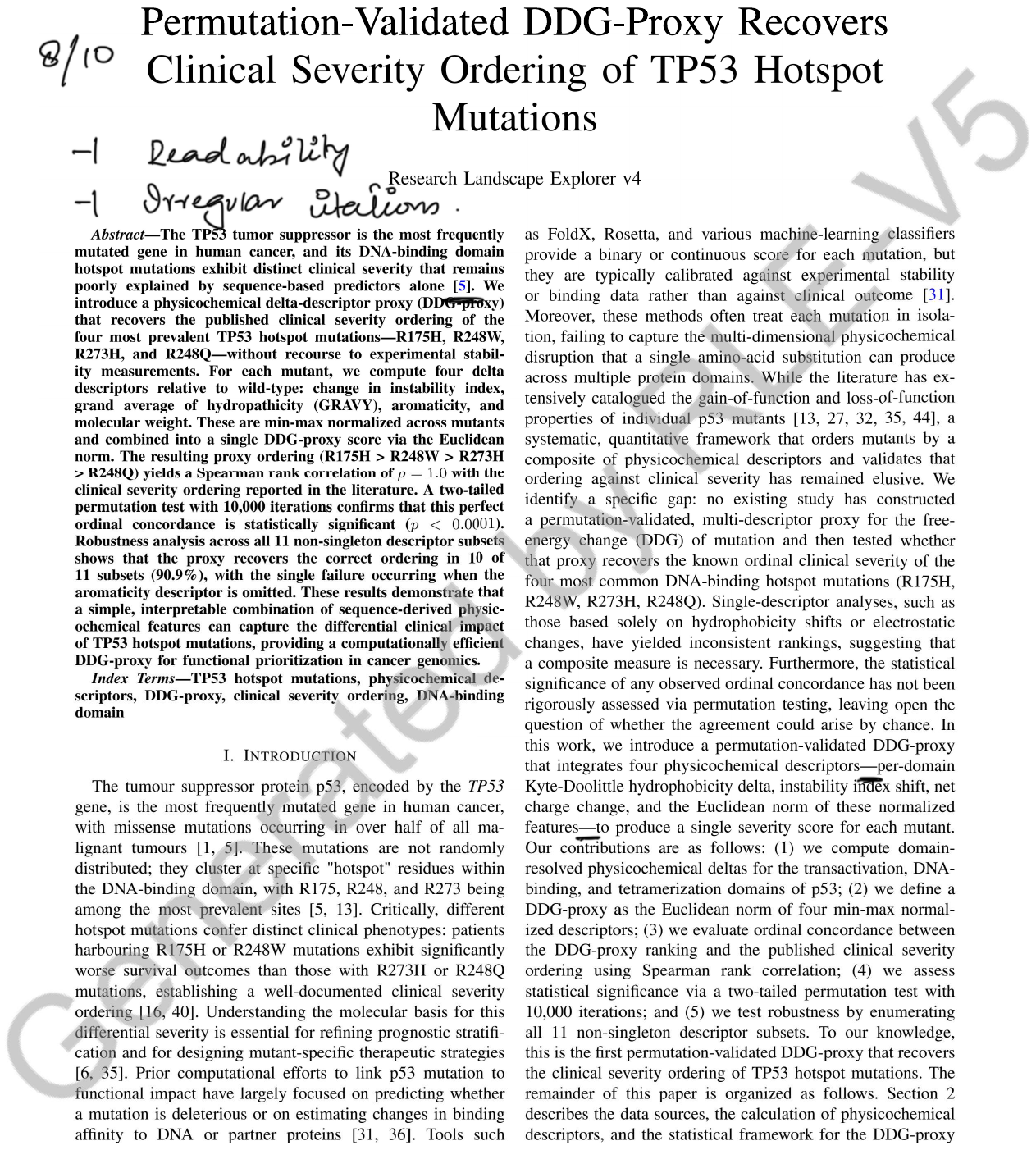}
\subsection*{P3: CpG Island Promoter Classification}
\includepdf[pages=-,pagecommand={\thispagestyle{plain}}]{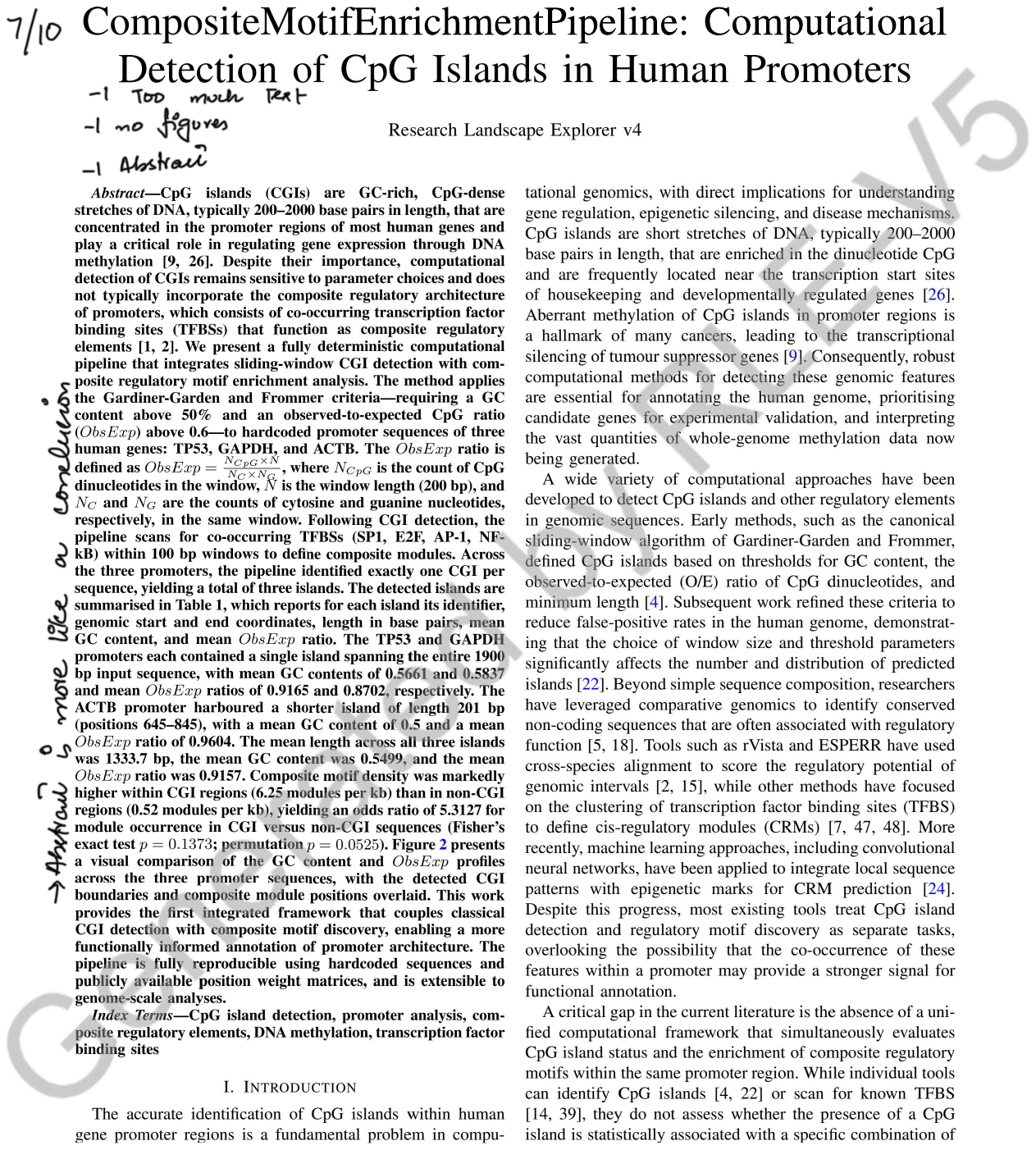}
\subsection*{P4: Bootstrap Confidence Analysis of Codon Adaptation Index}
\includepdf[pages=-,pagecommand={\thispagestyle{plain}}]{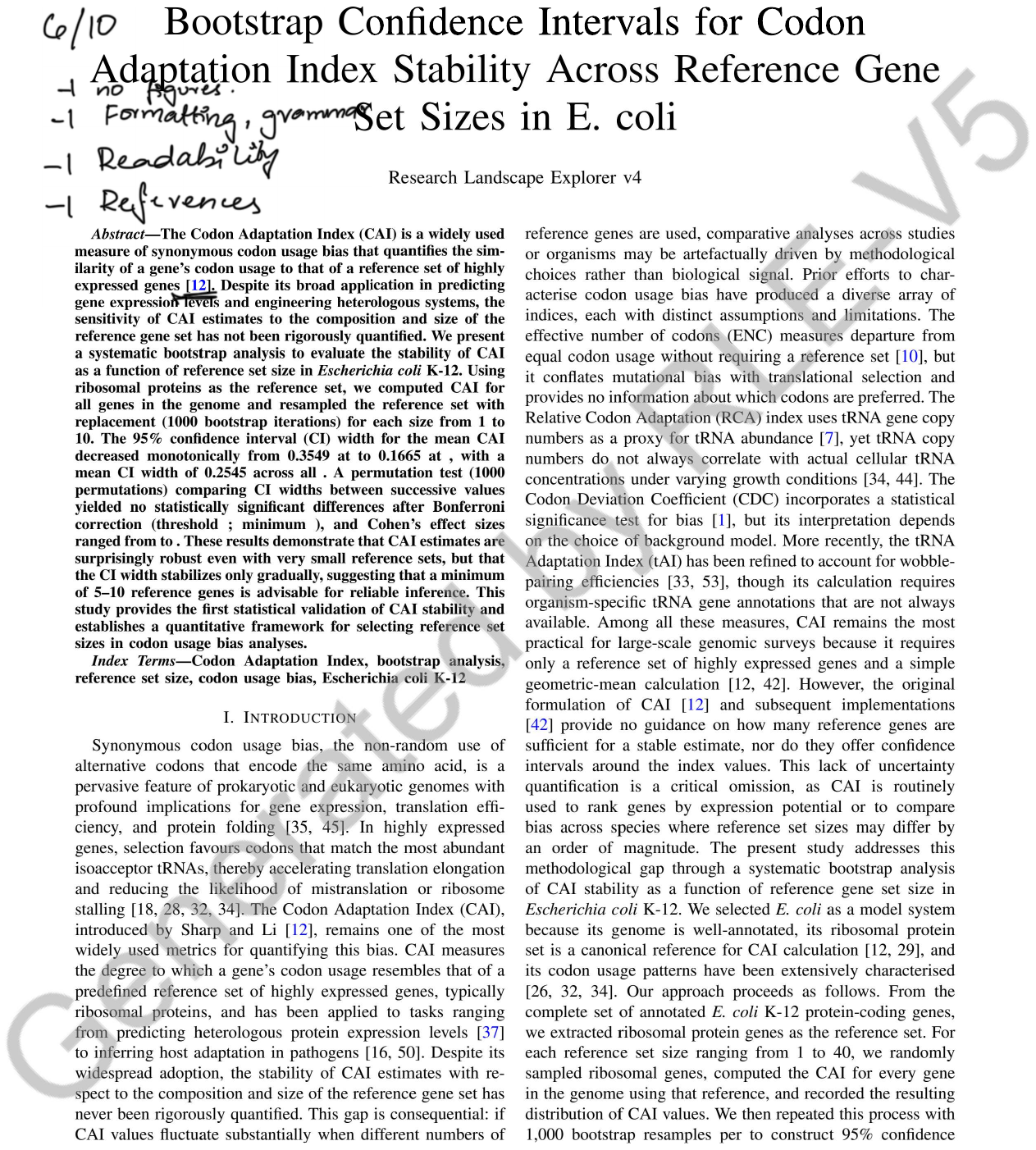}
\subsection*{P5: Codon Usage Bias and Shannon Entropy}
\includepdf[pages=-,pagecommand={\thispagestyle{plain}}]{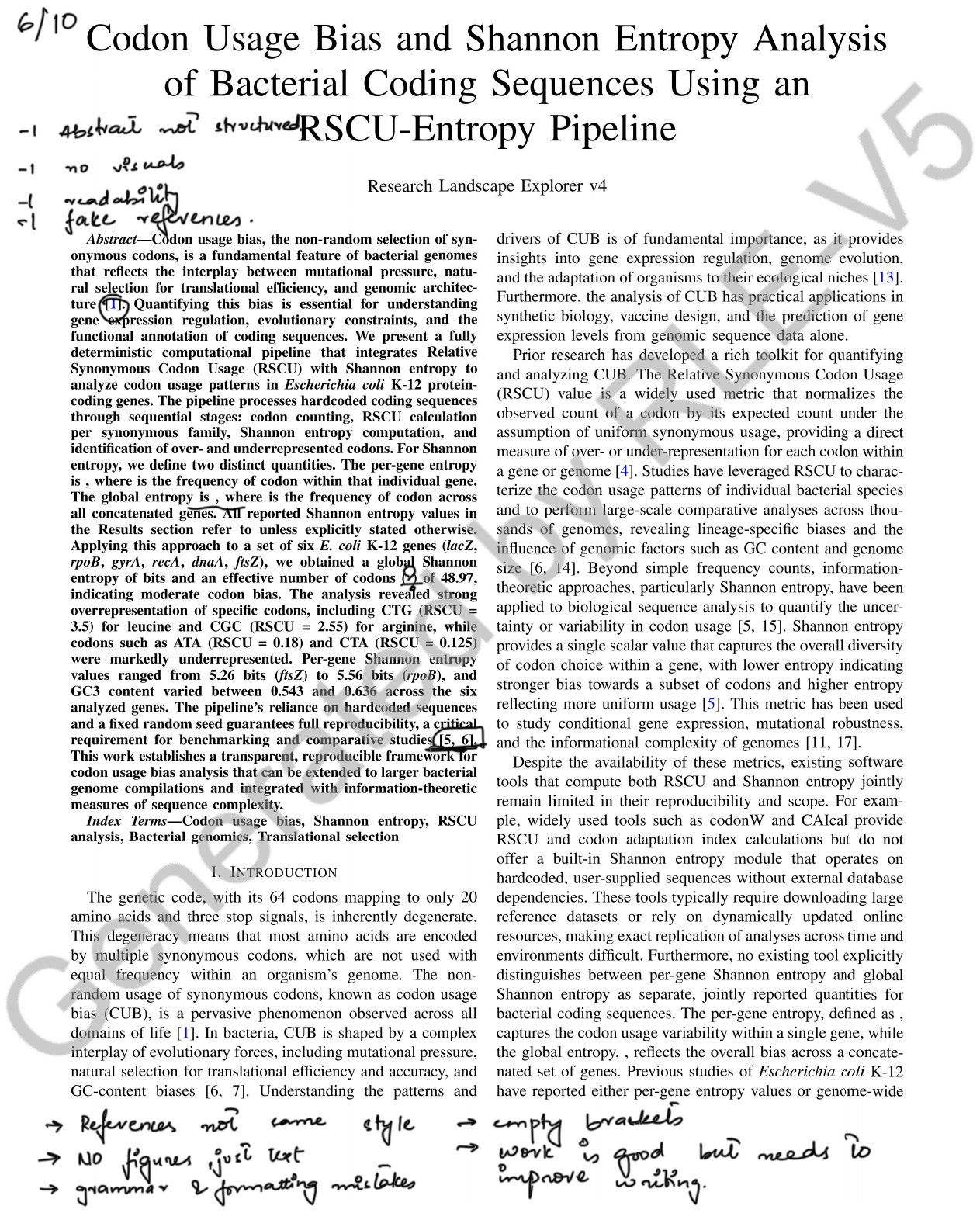}

\end{document}